\documentclass{article}

\newcommand{\imgpath}{./img}
\usepackage[preprint]{arxiv_neurips_2021}

\usepackage{times}
\usepackage{epsfig}
\usepackage{graphicx}
\usepackage{amsmath}
\usepackage{amssymb}
\usepackage{bbm}
\usepackage{authblk}
\usepackage[ruled,vlined]{algorithm2e}
\usepackage{footmisc}
\usepackage{mathtools}
\def\1{\boldsymbol{1}}

\def\va{{\boldsymbol{a}}}

\def\vd{{\boldsymbol{d}}}

\def\vg{{\boldsymbol{g}}}

\def\vm{{\boldsymbol{m}}}

\def\vq{{\boldsymbol{q}}}

\def\vs{{\boldsymbol{s}}}

\def\vx{{\boldsymbol{x}}}
\def\vy{{\boldsymbol{y}}}
\def\vz{{\boldsymbol{z}}}

\def\mU{{\boldsymbol{U}}}

\DeclareMathOperator{\E}{\mathbb{E}}

\usepackage{tabularx,booktabs}
\newcolumntype{Y}{>{\centering\arraybackslash}X}
\usepackage{caption} % Subfigures
\usepackage{subcaption} % Subfigures

\usepackage{hyperref}       % hyperlinks

\usepackage[capitalize,nameinlink]{cleveref}[0.19] % cref

\newcommand{\comment}[1]{}
\newcommand*\samethanks[1][\value{footnote}]{\footnotemark[#1]}
\title{ClimateGAN: Raising Climate Change Awareness by Generating Images of Floods}

\author[1,2]{Victor Schmidt\thanks{Corresponding authors - \texttt{[schmidtv, luccionis]@mila.quebec}}~~}
\author[1,2]{Alexandra Luccioni\samethanks~~}
\author[1,2]{Mélisande Teng}
\author[1,2]{Tianyu Zhang}
\author[1,2]{Alexia Reynaud}
\author[1, 3]{Sunand Raghupathi}
\author[1,2]{Gautier Cosne}
\author[1,2]{Adrien Juraver}
\author[4]{Vahe Vardanyan}
\author[1,2]{Alex Hernández-García}
\author[1,2]{Yoshua Bengio}
\affil[1]{Mila Québec AI Institute, Montréal, Canada}
\affil[2]{Université de Montréal, Montréal, Canada}
\affil[3]{Columbia University, New York City, USA}
\affil[4]{CDRIN, Matane, Canada}

\begin{document}

\maketitle

%%%%%%%%% ABSTRACT
\begin{abstract}
Climate change is a major threat to humanity, and the actions required to prevent its catastrophic consequences include changes in both policy-making and individual behaviour. However, taking action requires understanding the effects of climate change, even though they may seem abstract and distant. Projecting the potential consequences of extreme climate events such as flooding in familiar places can help make the abstract impacts of climate change more concrete and encourage action. As part of a larger initiative to build a website that projects extreme climate events onto user-chosen photos, we present our solution to simulate photo-realistic floods on authentic images. To address this complex task in the absence of suitable training data, we propose ClimateGAN, a model that leverages both simulated and real data for unsupervised domain adaptation and conditional image generation. In this paper, we describe the details of our framework, thoroughly evaluate components of our architecture and demonstrate that our model is capable of robustly generating photo-realistic flooding.
\end{abstract}

%%%%%%%%% BODY TEXT
\section{Introduction}
\label{sec:intro}

Climate change is an increasingly serious danger to our planet, with warming temperatures causing extreme weather events such as droughts, storms and heatwaves. These phenomena affect the livelihood of millions of people globally, with that number rising each year~\citep{hauer2017migration, neumann2015future, hoegh2018impacts,watts20192019}. Climate change is also making flooding worse, both due to rising sea levels as well as increasing precipitation and faster snow melt in some areas, presenting a major risk to both coastal and in-land populations worldwide~\citep{dottori_development_2016, vousdoukas_global_2018}.

Visualizing the effects of climate change has been found to help overcome \emph{distancing}, a psychological phenomenon resulting in climate change being perceived as temporally and spatially distant and uncertain~\citep{sheppard2012visualizing, spence2012psychological}, and thus less likely to trigger action. In fact, images of extreme weather events~\citep{leviston2014imagining} and their impacts~\citep{chapman2016climate} have been found to be especially likely to trigger behavioral changes. 

Previous research has shown that simulating first-person perspectives of climate change can contribute to reducing distancing~\citep{berenguer2007effect, sevillano2007perspective}. Digital technologies are increasingly used for this purpose, ranging from interactive data dashboards~\citep{herring2017communicating} to web-based geographic visualizations~\citep{neset2016climate, glaas2017visualization} to immersive digital environments of ecosystems~\citep{ahn2014short}. However, existing technologies have targeted specific environments or regions, manually rendering climate change effects; there is, to our knowledge, no tool that can render impacts of climate change in arbitrary locations to that end.

\begin{figure}
\centering
\includegraphics[width=\textwidth]{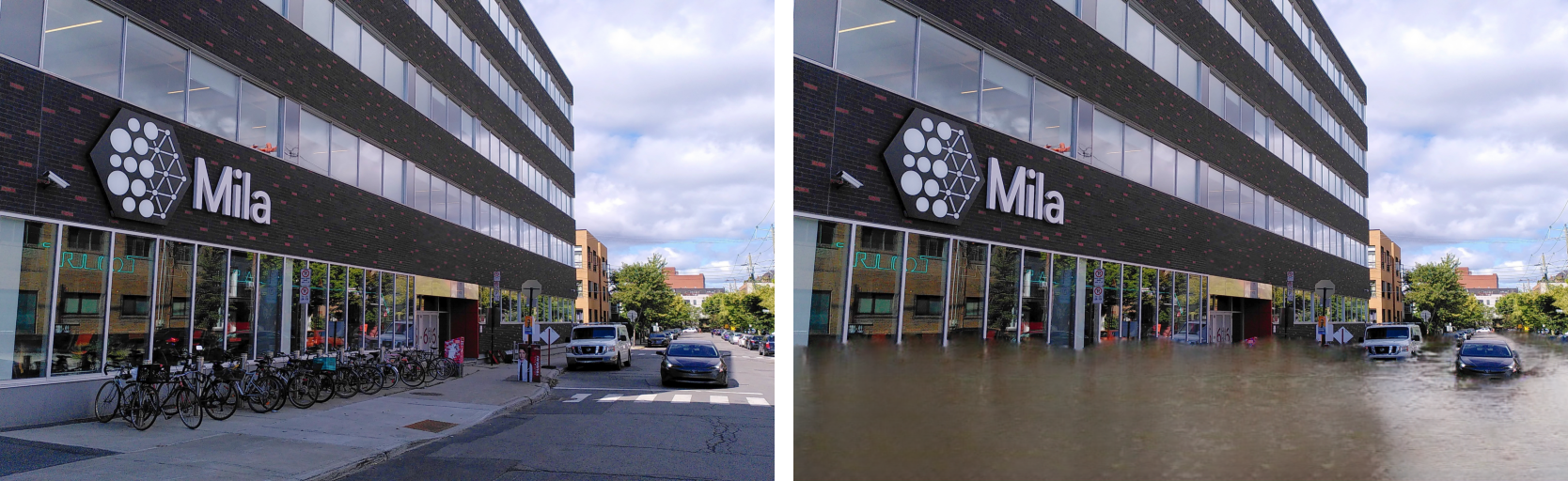}
\caption{\textit{We present ClimateGAN, a model that simulates extreme floods (right) on real scene images (left).}}
\label{fig:intro}
\end{figure}

In this context, we endeavor to develop a system enabling users to enter an address of their choice and visualize the associated first-person image from Google Street View, transformed as if it had been impacted by extreme flooding. To do so, we propose our model \emph{ClimateGAN}, which decomposes the task of flooding into two parts: a \emph{Masker} model to predict which pixel locations in an image would be under water if a flood occurred, and a \emph{Painter} model to generate contextualized water textures conditionally on both the input image and the Masker's prediction. Our contributions are: proposing and motivating the new task of generating floods, a novel architecture for the Masker, a data set of images of floods from a virtual world (\cref{sec:creating-images-of-floods}), and a procedure to evaluate such a model in the absence of ground-truth data (\cref{sec:evaluation-method}). We also compare our model to existing generative modeling frameworks and provide a thorough ablation study of the components of our model (\cref{sec:results}). We conclude with future work and ways of improving and extending our tool, as well as our plans to deploy it on a website to raise climate change awareness (\cref{sec:limitations-and-improvements,sec:conclusion}).

\section{Related Work} 
\label{sec:relatedwork}
 
While the task of generating extreme street-level flooding in images retrieved from Google StreetView is novel, related work has been carried out in applying Deep Learning for flood segmentation~\citep{sazara2019detecting} and flood depth estimation~\citep{kharazi2021flood}. Generative modeling has also been used for transferring weather conditions on street scenes using both imagery~\citep{li2021weather} and semantic maps~\citep{wenzel2018modular}. For the purposes of our task and given its unique constraints, we frame our approach in the context of {\em image-to-image translation}, involving {\em conditional image synthesis} and {\em domain adaptation}. We present relevant related work from these three areas in the sections below.

\paragraph{Image-to-image translation}
(IIT) is a computer vision task whose goal is to map a given image from one domain to another~\citep{liu2017nsupervised, huang_multimodal_2018, yi2017dualgan}. We can divide IIT approaches into two categories: those that carry out the translation on the entire input image, and those that utilize masks to guide the translation task. In the first category, several notable architectures have enabled the field to progress in new directions. While IIT initially relied on the existence of two aligned domains such as photographs and sketches of the same objects, CycleGAN~\citep{zhu_unpaired_2017} overcame this constraint, learning two models and regularizing them via a \textit{cycle-consistency loss} that allowed for the preservation of the symmetry of the transformation while allowing the domains themselves to remain unaligned. Further progress in IIT was made by architectures such as MUNIT~\citep{huang_multimodal_2018}, which enabled the disentanglement of content and style, allowing multimodal translations, FUNIT~\citep{liu2019few}, which enabled few-shot IIT, and, more recently, the CUT model~\citep{park2020contrastive}, which enabled single image unpaired translation with the help of contrastive learning. 

A second category of IIT focuses the translation process on particular input image areas, typically by leveraging attention or segmentation masks. This can be useful in more targeted translation, for instance when only certain objects in the foreground of an input image are meant to be translated. This more closely resembles our case, as we aim to transform only one part of the image with flooding water. Examples in this category include Attention-Guided GANs~\citep{tang2019attention}, which learn attention masks during training and utilize them at inference time to render more precise transformations, and InstaGAN~\citep{mo_instagan_2018} which uses instance-level semantic masks to guide the translation process.

\paragraph{Conditional image synthesis} differs from IIT in that the input can be a label, text or a segmentation map, instead of another image~\citep{mirza2014conditional, frolov2021adversarial, Hong_2018_CVPR, zhu2020sean}. One approach from this category that is particularly relevant to our work is SPADE (SPatially-Adaptive (DE)normalization)~\citep{park_semantic_2019}, a module that enables the transformation of a semantic layout---such as that of a street scene or landscape---into an image that semantically matches this layout. This approach also introduced the GauGAN generator, which leverages SPADE blocks to learn a spatially-adaptive transformation, enabling the synthesis of realistic images based on the input maps. 

\paragraph{Domain adaptation}
is a branch of transfer learning that aims at transferring knowledge from one domain to another using different data sources~\citep{pan_survey_2010, ganin_unsupervised_2015}. This can be particularly useful in tasks where more (labeled) data is available from a simulated world than in the real world, like in our application. Domain adaptation techniques can then be used to bridge the distributional gap between real and simulated scenes, learning useful tasks such as semantic segmentation and depth prediction, which function both in real and simulated scenarios. Examples of domain adaptation approaches that adopt these techniques include: CYCADA~\citep{hoffman_cycada_2017}, which leverages cycle-consistency constraints similar to those proposed by CycleGAN~\citep{zhu_unpaired_2017} to improve domain adaptation, ADVENT~\citep{vu_advent_2018}, which uses Adversarial Entropy Minimization to achieve state-of-the-art performance in unsupervised domain adaptation for semantic segmentation, and Depth-aware Domain Adaptation (DADA)~\citep{vu_dada_2019}, which improves on ADVENT by leveraging dense depth maps. 
%cite dhamo2019peeking here? I really don't think it brings much to the table

\section{Creating Images of Floods}
\label{sec:creating-images-of-floods}

Our task of generating flooding resembles that of unsupervised image-to-image translation. However, this framework poses two major problems: first, the translation needs to be restricted to the portion of the image that would contain water, so we cannot consider approaches which alter the image globally. Second, we are only concerned with adding water and not the reverse, which eliminates approaches that rely on the symmetry of the translation task. In order to undertake this task, we therefore developed a novel conditional image synthesis approach that consists of two models: a Masker that produces a binary mask of where water would plausibly go in the case of a flood, and a Painter that renders realistic water given a mask and an image. We provide an overview of this procedure in \cref{fig:overview}, and detail the individual components in this section.

\subsection{Data} 
\label{sec:data}

To overcome the lack of available data, we combined real images and simulated data from a virtual world, binding the two sources using unsupervised domain adaptation.

\subsubsection{Real Data}

First-person images of floods are scarce; moreover, even when such pictures exist, it is exceedingly rare to find the corresponding image of the same location before the flooding. We therefore pursued several alternate approaches for gathering real flooded images, spanning from web-scraping to creating a website and a mobile app %\footnote{URL hidden for the review process.}
to collect crowd-sourced images. We complemented flooded images with images of `before' floods, i.e. of normal streets and houses. We aimed to cover a broad scope of geographical regions and types of scenery: urban, suburban and rural, with an emphasis on images from the Cityscapes~\citep{cordts_cityscapes_2016} and Mapillary~\citep{neuhold2017mapillary} data sets, which closely resemble the viewpoint of Google Street View. We collected a total of 6740 images: 5540 non-flooded scenes to train the Masker, and 1200 flooded images to train the Painter.

\subsubsection{Simulated Data}
\label{sec:simdata}
Besides the lack of paired data, another limitation of real-world images is that they do not contain scene geometry annotations and semantic segmentation labels, which we want to leverage during model training. To solve both of these problems, we created a 1.5 $\textit{km}^2$ virtual world using the Unity3D game engine. To be as realistic as possible, it contains urban, suburban and rural areas, which we flooded with approximately 1m of water to gather `before' and `after' pairs (see~\cref{fig:sim} in the supplementary materials, for example images). For each pair of images obtained, we also captured the corresponding depth map and semantic segmentation layout of the scene.
%Finally, to be as close as possible to real-world testing conditions, that is, images from Google Street View, we parameterized the virtual camera settings, such as field of view, height and pitch, correspondingly. 
In the end, we gathered approximately 20000 images from 2000 different viewpoints in the simulated world, which we used to train the Masker. We make this data set publicly available\footnote{https://github.com/cc-ai/mila-simulated-floods} to enable further research.

\subsection{Masker}
\label{sec:masker}

While we have access to ground-truth supervision in the simulated domain, we need to transfer the Masker's knowledge to real images to perform our task. We therefore adapted the ADVENT methodology~\citep{vu_advent_2018} to train the Masker, providing additional depth information to improve its predictions as proposed by the authors of DADA~\citep{vu_dada_2019}. Moreover, we introduced a segmentation decoder to inform the Masker of the semantics of the input image, since we hypothesized that adding this information to the flooding process would help the Masker produce more sensible flood masks. To that end, we structured the Masker as a multi-headed network with a shared encoder and three decoders performing depth prediction, semantic segmentation and masking. We trained these decoders simultaneously in a multi-task way by minimizing the sum of their losses. We provide further details on the losses used by the Masker in subsequent sections, as well as in~\cref{sup:losses}.

\begin{figure}[t]
\begin{center}
\includegraphics[width=0.65\textwidth]{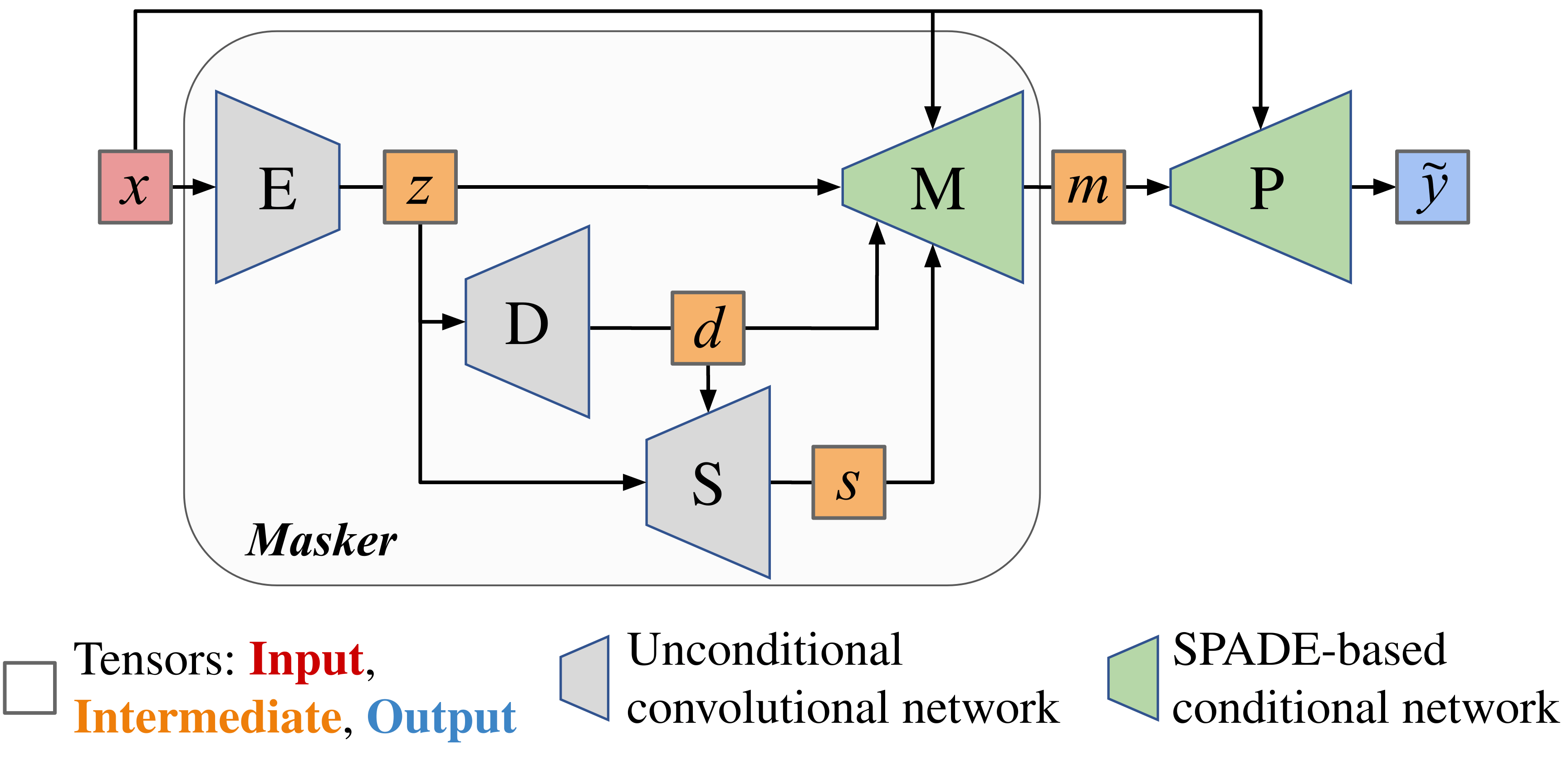}
\end{center}
\caption{The ClimateGAN generation process: first, the input $\vx$ goes through the shared encoder $E$. Three decoders use the resulting representation $\vz$: $D$ predicts a depth map $\vd$, $S$ produces a depth-informed segmentation map $\vs$, and lastly $M$ outputs a binary flood~mask, taking $\vz$ as input and sequentially denormalizing it with SPADE blocks~\citep{park_semantic_2019}, conditionally on $\vd$, $\vs$ and $\vx$. Finally, the SPADE-based Painter $P$ generates an image $\tilde\vy$ of a flood, conditioned on the input image $\vx$ and the binary predicted mask $\vm$. Note that the Masker and the Painter are trained independently and only combined at test-time.}
\label{fig:overview}
\end{figure}

In the following section, subscripts $s$ and $r$ respectively identify the simulated and real domains; we use $i\in \{r, s\}$ to refer to an arbitrary domain. $E$ is an encoder network while $D$, $S$ and $M$, are the depth, segmentation and flood mask decoders, respectively, as per~\cref{fig:overview}.

To guide decoders in the early training stages, we used \textit{pseudo labels} as supervision targets. In other words, we considered predictions of \textit{ad hoc} pretrained models as ground truth. However, because those labels are noisy, we limited that procedure to the beginning of training.

\paragraph{Depth decoder}
We consider depth in the disparity space, and predict the normalized inverse depth $\vd_i~=~D(E(\vx_i))$ from an input image $\vx_i$. We used the scale-invariant loss from MiDaS~\citep{midas}, which is composed of a scale and shift-invariant MSE loss term $\mathcal{L}_{SSIMSE}$ and a gradient matching term $\mathcal{L}_{GM}$. We used the following targets to compute this loss: for simulated input images, we leveraged available ground-truth depth maps, and for real input images, we used pseudo labels inferred from the MiDaS v2.1 model~\citep{midas}.
The complete depth loss is: 
\begin{equation}
\begin{split}
  \mathcal{L}_{Depth} &= \lambda_{1}\mathcal{L}_{SSIMSE} + \lambda_{2} \mathcal{L}_{GM}.
\end{split}
\end{equation}

\paragraph{Segmentation decoder}
The segmentation decoder $S$ is implemented such that $S\circ E$ corresponds to the DeepLabv3+ architecture~\citep{deeplabv3plus2018}. It is trained as described in~\citep{vu_dada_2019}, leveraging depth information available in the simulated world to improve segmentation predictions by giving more attention to closer objects, producing $\vs_i=S(E(\vx_i) ,\vd_i)$. Two fusion mechanisms encourage this: \textit{feature fusion}, which multiplies element-wise the latent vector $\vz_i=E(\vx_i)$ by a depth vector obtained from the depth decoder, and \textit{DADA fusion}, which multiplies the {\em self-information map} $I(\vs_i)$ element-wise with the depth predictions $\vd_i$ to obtain the depth-aware self-information map $\hat{I}(\vs_i)~=~I(\vs_i)~\odot~\vd_i$. In this context, the self-information map of a probability distribution $\vq_i$ is defined~\citep{vu_advent_2018} as $I(\vq_i)^{(h,w)}=-\vq_i^{(h,w)}\cdot~\log \vq_i^{(h,w)}$, where $\vq_i^{(h,w)}$ is a probability distribution over channels (\textit{i.e.} semantic classes for $\vs_i$) at a given location $(h, w)$.

The training procedure of the segmentation decoder starts with the cross-entropy loss $\mathcal{L}_{CE}(\vy_i, \vs_i)$ between predictions $\vs_i$ and labels $\vy_i$. In the simulated domain, these labels correspond to the ground truth. In contrast, in the real domain, we used pseudo labels inferred from HRNet~\citep{tao2020ierarchical}, a model that has achieved state-of-the art results in semantic segmentation on the Cityscapes data set~\citep{cordts_cityscapes_2016}.
A key difference between the real and simulated domains is the model's confidence in its predictions. Therefore, to encourage real domain predictions to be confident and reduce the gap with simulated predictions, an entropy minimization (EM) term $\mathcal{L}_{EM}(\vs_r)$ similar to~\cref{eq:EM} is added.

Finally, we shrank the domain gap between the distributions of real and simulated self-information maps using adversarial training.
A discriminator $Q^S$ is introduced to distinguish between the two depth-aware maps and is trained using a WGAN objective~\citep{arjovsky2017wasserstein}:
$\mathcal{L}_{WGAN}(Q^S,\hat{I}(\vs_r),\hat{I}(\vs_s))$.

The overall loss of the segmentation decoder is therefore:
\begin{equation}
\begin{split}
  \mathcal{L}_{Seg} &= \lambda_{3} \mathcal{L}_{CE} + \lambda_{4} \mathcal{L}_{EM} + \lambda_{5}\mathcal{L}_{WGAN}.
\end{split}
\end{equation}

\paragraph{Flood mask decoder}
This decoder is structured to be conditioned not only on the input image, but also on predictions $\vd_i$ and $\vs_i$ from other decoders. To implement this dependence, we used SPADE~\citep{park_semantic_2019} conditional blocks. The idea behind SPADE is to create residual blocks where the input is first normalized and then denormalized in a spatially relevant way by small convolutional networks $\gamma$ and $\beta$, functions of the conditioning variables $\mU$. Given an activation $\va_{n, c, h, w}$ at a given layer, its mean per channel $\mu_c$ and standard deviation per channel $\sigma_c$, the output of the SPADE denormalization layer is:
\begin{equation}
  \gamma_{c, h, w}(\mU) \frac{\va_{n, c, h, w} - \mu_c}{\sigma_c} + \beta_{c, h, w}(\mU).
\end{equation}

In our case, for an input $\vx_i$, the conditioning variable is therefore $\mU_i=[\vx_i, \vd_i, \vs_i]$, where we concatenate the tensors along the channel axis. The mask $\vm_i = M(\vz_i, \mU_i)$ and its self-information map $I(\vm_i)$ are computed from the latent representation $\vz_i = E(\vx_i)$. We also included a total variation (TV) loss on the mask $\vm_i$ for both domains in order to encourage the predictions to be smooth, ensuring that neighboring pixels have similar values~\citep{johnson2016erceptual}---note that $\Delta$ here means spatial difference of the image mesh.
\begin{equation}
\begin{split}
  \mathcal{L}_{TV}(\vm_i) &= \E_{n,h,w}[ (\Delta_{h} \vm_i)^2 + (\Delta_{w} \vm_i)^2 ].
\end{split}
\end{equation}

In the simulated domain, we used a standard binary cross-entropy loss $\mathcal{L}_{BCE}(\vy_{m_s},\vm_s)$ with the ground-truth mask $\vy_{m_s}$. In the real domain, where we have no ground truth, we encouraged the predicted flood mask $\vm_r$ to at least encompass the ground by introducing a ground intersection (GI) loss, penalizing masks that assign a low probability to locations where a pre-trained model detected ground. The formulation of this loss leverages pseudo labels $\vg_r$ from HRNet~\citep{tao2020ierarchical} for the ground mask in $\vx_r$:
\begin{equation}
  \mathcal{L}_{GI}(\vg_r, \vm_r) = \E_{n,h,w}[ \mathbbm{1}_{(\vg_r - \vm_r) > 0.5}].
\end{equation}

As per the ADVENT approach, we also added an entropy minimization loss to increase the mask decoder's confidence in its real domain predictions:
\begin{equation}\label{eq:EM}
\begin{split}
  \mathcal{L}_{EM} (\vm_r)= \E_{n,c, h, w} [-\vm_r \log \vm_r].
\end{split}
\end{equation}
Lastly, similarly to the segmentation decoder, we adversarially trained the flood mask decoder with the loss $\mathcal{L}_{WGAN}(Q^M, I(\vm_r), I(\vm_s))$ to attempt to produce self-information maps $I(\vm_i)$ indistinguishable by a discriminator $Q^M$. All in all, $M$'s total loss is a weighted sum of all of the above losses:
\begin{equation}
\begin{split}
  \mathcal{L}_{Mask} = \lambda_{6} \mathcal{L}_{TV} + \lambda_{7} \mathcal{L}_{GI} + \lambda_{8} \mathcal{L}_{BCE} + \lambda_{9}\mathcal{L}_{EM} + \lambda_{10}\mathcal{L}_{WGAN}. 
\end{split}
\end{equation}

The Masker's final loss sums the losses of the three decoders: $\mathcal{L}_{Masker} = \mathcal{L}_{Depth} + \mathcal{L}_{Seg} + \mathcal{L}_{Mask}$.

\subsection{Painter}

Given an input image and a binary mask, the goal of the Painter is to generate water in the masked area while accounting for context in the input image. This is important for the realism of a flooded image because water typically reflects the sky and surrounding objects, and is colored conditionally on the environment. We built our Painter model based on the GauGAN~\citep{park_semantic_2019} architecture, using SPADE conditioning blocks. As we explained in the previous section, these blocks learn transformations that vary both spatially and conditionally on the input, allowing the model to better propagate spatial semantic information through the generation pipeline than if it were only available at the input of the generator. We adapted GauGAN to fit our task: rather than conditioning on a semantic map, we conditioned on a masked image. 
We trained the Painter on the 1200 real flooded images and pseudo labels of water segmented by a trained DeepLabv3+ model~\citep{chen2017deeplab}, whereas at inference time, the mask $\vm$ for an input image $\vx$ is generated by the Masker. Finally, we copied the context (non-masked area) back onto the generated image, to ensure that the other objects in the image (e.g. buildings and sky) remain intact. Thus, the output $\tilde\vy$ of the Painter $P$ is:
\begin{equation}
\label{eq:copy-paste}
  \tilde\vy = P(\epsilon, (1 - \vm) \odot \vx ) \odot \vm + \vx \odot (1-\vm), 
\end{equation}
where $\epsilon \sim N(\textbf{0}, \textbf{I})$. The Painter generator is a sequence of 7 SPADE-denormalized residual blocks followed by bilinear upsampling operations with a scale factor of 2. At each level, an interpolation of the masked input image with the appropriate size is given as conditioning variables $\mU$ to the corresponding SPADE block. We used a multi-resolution PatchGAN discriminator, as was done in the original GauGAN implementation~\citep{isola_image--image_2017}.
According to this paper, a perceptual VGG loss~\citep{ledig_photo-realistic_2016} and a discriminator feature-matching loss~\citep{salimans2016improved} are essential for good performance. Since these last two losses rely on paired examples, we trained the Painter separately from the Masker using images of floods.

\begin{figure}
\centering
\small
\begin{tabularx}{\textwidth}{*{6}{Y}}
Input & Depth & Segmentation & Mask & Masked Input & Painted Input\\
\end{tabularx}
\includegraphics[width=\textwidth]{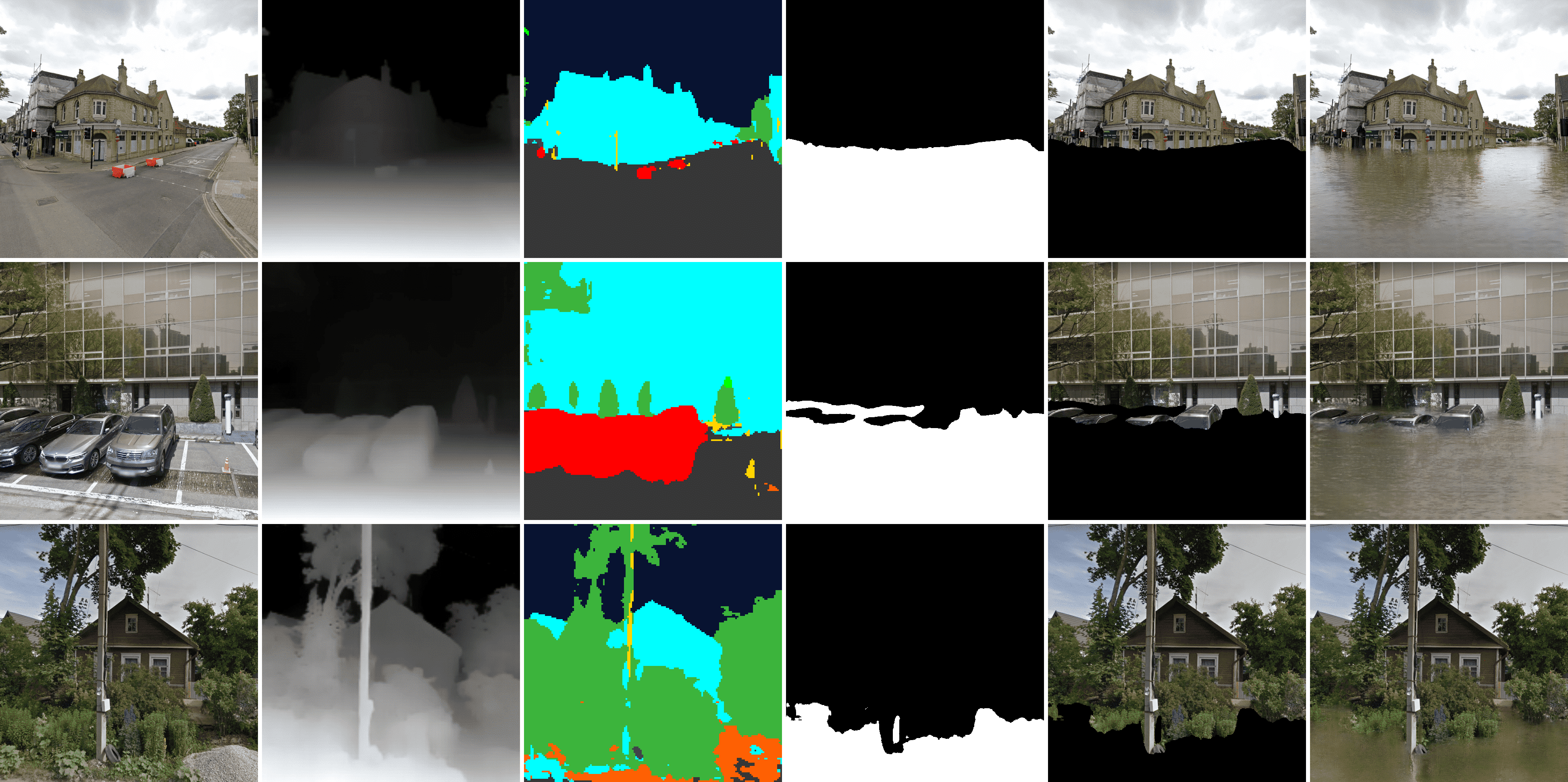}
\caption{Example inferences of ClimateGAN, along with intermediate outputs. The first row shows how the Masker is able to capture complex perspectives and how the Painter is able to realistically contextualize the water with the appropriate sky and building reflections. On the second row, we can see that close-ups with distorted objects do not prevent the Masker from appropriately contouring objects. Finally, the last row illustrates how, in unusual scenes, the Masker may imperfectly capture the exact geometry of the scene but the final rendering by the Painter produces an acceptable image of a flood. More inferences including failure cases are shown in \cref{fig:more-climategan}.}
\label{fig:climategan-inferences}
\end{figure}

\section{Evaluation Method}
\label{sec:evaluation-method}

The quality and photo-realism of our model's output depend on both the accuracy of the Masker in determining a realistic area for the flood and the ability of the Painter to infill the input mask with water with a realistic texture matching the surrounding scene. In order to best understand the contribution of each component, we evaluated the Masker and the final flooded images separately.

\subsection{Masker Evaluation}
\label{sec:masker-evaluation}

The lack of ground-truth data is not only an obstacle for training, but also for obtaining reliable evaluation metrics to assess the performance of our model. Therefore, in order to obtain metrics for evaluating the quality of the Masker and for comparing the individual contribution of the proposed components to its architecture, we manually labeled a test set of 180 images retrieved from Google Street View. We collected images of diverse geographical provenance, level of urban development, and composition (crowds, vehicles, vegetation types, etc.). We manually annotated every pixel of each image with one of three classes: (1) \emph{cannot-be-flooded}---pixels higher than 1.5m above the ground level; (2) \emph{must-be-flooded}---anything with height less than 0.5 m, and (3) \emph{may-be-flooded}---all remaining pixels. We provide further details in~\cref{sup:test-annotations}.

\subsubsection{Metrics}
\label{sec:metrics}
We propose the following metrics to evaluate the quality of masks (more details in \cref{sec:suppl:metrics}):

\paragraph{Error rate} 
The perceptual quality of the final images is highly impacted by the area of the image with errors in the predicted mask. We found that both large gaps in \textit{must-be-flooded} areas (i.e. false negatives, $FN$), and large sections of predicted masks in \textit{cannot-be-flooded} areas (i.e. false positives, $FP$) account for low perceptual quality. Thus, we propose the error rate to account for the amount of prediction errors in relation to the image size:
\begin{equation}
\label{eq:error}
  error = (FN + FP) / (H \times W).
\end{equation}

\paragraph{F05 Score}
So as to consider the precision and recall of the mask predictions, we also evaluated the $F_{\beta=0.5}$ (F05) score, which lends more weight to precision than to recall~\citep{van1979information}.

\paragraph{Edge coherence}
It is also important to take into account the 
\textit{shape similarity} between the predicted mask and the ground-truth label, particularly in the uncertainty region defined by the \textit{may-be-flooded} class. We capture this property by computing the standard deviation $\sigma$ of the minimum Euclidean distance $d(\cdot, \cdot)$ between every pixel in the boundary of the predicted mask $\hat{B}$ and the boundary of the \textit{must-be-flooded} area $B$, which are computed by filtering the masks with the Sobel edge detector:
\begin{equation}
\label{eq:edge-coherence}
  edge\ coherence = 1 - \sigma \left( \min_j \left[ d(\hat{B_i}, B_j)/H \right] \right).
\end{equation}

\subsubsection{Ablation Study}
\label{sec:ablation-method}

In order to assess the contribution of each of the components described in~\cref{sec:masker} to the overall Masker performance, we performed an ablation study by training 18 models, each with a different combination of techniques. As illustrated in ~\cref{tab:ablation}, we analyzed the effect of including the following components in the Masker architecture: training with pseudo labels, $D$, $S$, DADA for $S$~\citep{vu_dada_2019}, DADA for $M$, and the SPADE~\citep{park_semantic_2019} architecture for $M$. When not using SPADE, $M$ is based on residual blocks followed by convolutional and upsampling layers, taking $z$ as input, and we can also inform it with depth according to the DADA procedure just like we do for $S$.

\begin{table}[b]
\centering
\resizebox{\columnwidth}{!}{%
\begin{tabular}{@{}lcccccccccccccccccccccc@{}}
\toprule
                   & 1 & 2 & 3 & 4 & 5 & 6 & 7 & 8 & 9 & 10 & 11 & 12 & 13 & 14 & 15 & 16 & 17 & 18 & G & I \\ \midrule
Pseudo labels      & • & • & • & • & • & • & • & • & • &    &    &    &    &    &    &    &    &    &   &   \\
Depth ($D$)        &   & • &   & • & • & • & • & • & • &    & •  &    & •  & •  & •  & •  & •  & •  &   &   \\
Segmentation ($S$) &   &   & • & • & • & • & • & • & • &    &    & •  & •  & •  & •  & •  & •  & •  &   &   \\
SPADE              &   &   &   &   & • &   & • &   &   &    &    &    &    & •  &    & •  &    &    &   &   \\
DADA ($S$)         &   &   &   &   &   & • & • &   & • &    &    &    &    &    & •  & •  &    & •  &   &   \\
DADA ($M$)         &   &   &   &   &   &   &   & • & • &    &    &    &    &    &    &    & •  & •  &   &   \\ \bottomrule\\
\end{tabular}
}
\caption{Summary of the ablation study of the Masker. Each numbered column corresponds to a trained model and black dots indicate which techniques (rows) were included in the model. The last two columns correspond to the baseline models: ground (G) segmentation from HRNet as flood mask and InstaGAN (I).}
\label{tab:ablation}
\end{table}

We compared these variants of the Masker module with two baselines: ground segmentation from HRNet~\citep{tao2020ierarchical} instead of the flood mask (G) and InstaGAN (I)~\citep{mo_instagan_2018}. For each model variant in the ablation study (columns) we computed mask predictions for each of the 180 test set images and the values of the three metrics: error, F05 and edge coherence. In order to determine whether each technique (row) improves the performance, we computed the difference between the metrics on each image for every pair of models where the only difference is the inclusion or exclusion of this technique. Finally, we carried out statistical inference through the percentile bootstrap method~\citep{efron1992bootstrap, rousselet2019bootstrap} to obtain robust estimates of the performance differences and confidence intervals. In particular, we obtained one million bootstrap samples (sampling with replacement) for each metric to get a distribution of the bootstrapped 20~\% trimmed mean, which is a robust measure of location~\citep{wilcox2011hypothesistesting}. We then compared the distribution against the null hypothesis, which indicates no difference in performance (see ~\cref{fig:ablation-bootstrap-ci}). We considered a technique to be beneficial for the Masker if its inclusion reduced the error rate. In case of inconclusive error rate results, we considered an increase in the F05 score and finally in the edge coherence.

\begin{figure}[th]
\begin{tabularx}{\textwidth}{*{7}{Y}}
\small{Input} & \small{CycleGAN} & \small{MUNIT} & \small{InstaGAN} & \small{InstaGAN+Mask} & \small{Painter+Ground} & \small{ClimateGAN}
\end{tabularx}
\centering
\includegraphics[width=\textwidth]{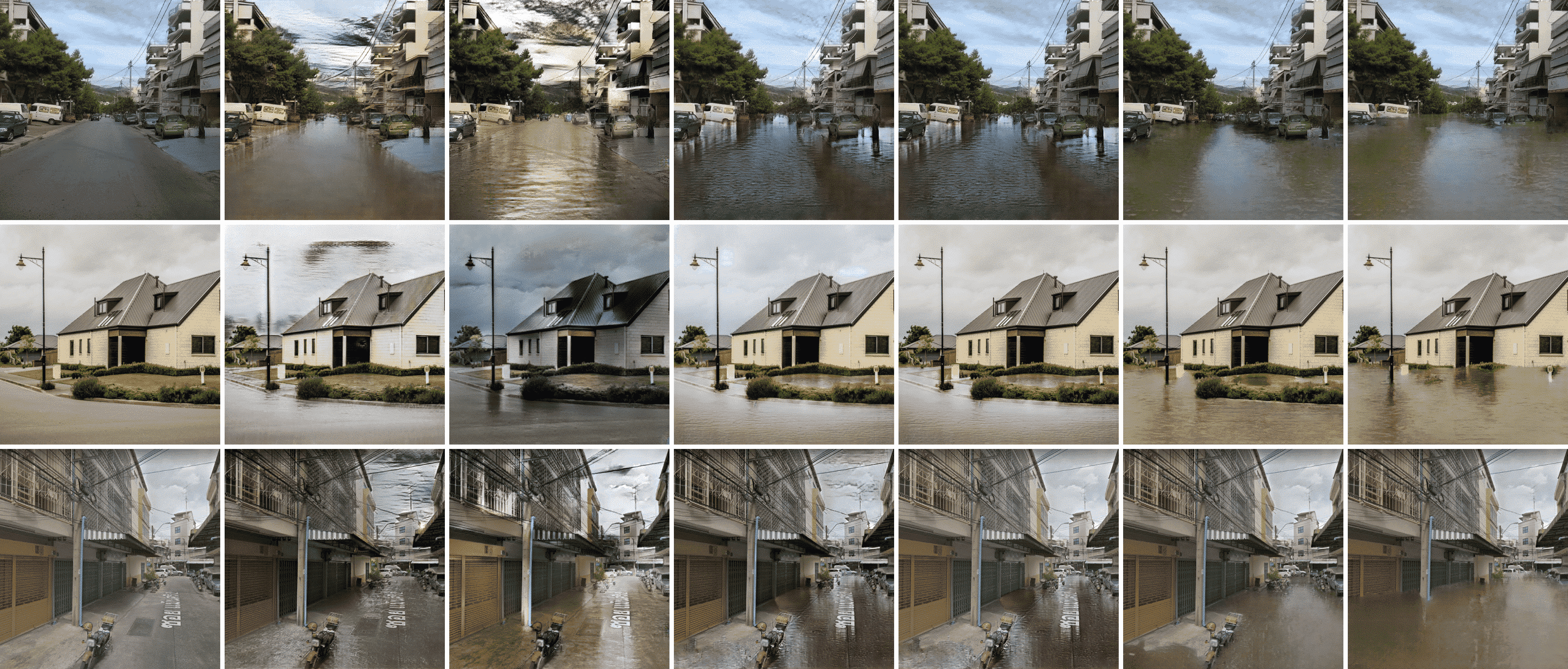}
\caption{Example inferences of ClimateGAN and comparable approaches on three diverse street scenes from the test set. We can see that ClimateGAN is able to generate both realistic water texture and color, as well as a complex mask that surrounds objects such as cars and buildings. Comparable approaches are often \emph{too destructive}, producing artifacts in the buildings and sky (e.g. $1^{st}$ row of MUNIT) or \emph{not destructive enough}, resembling more rain on the ground than high floods (e.g. $3^{rd}$ row of CycleGAN and $2^{nd}$ row of InstaGAN).}
\label{fig:comparables}
\end{figure}

\subsection{Comparables} 

While the nature of our task is specific to our project, we can nonetheless benchmark ClimateGAN against IIT models.
In fact, in earlier iterations of our project, we leveraged the CycleGAN~\citep{zhu_unpaired_2017} architecture in order to achieve initial results~\citep{vicc2019}, before adopting a more structured approach. Therefore, to be as comprehensive in our benchmarking as possible, we trained the following five models on the same data as the ClimateGAN Painter and used the same test set for the comparison: CycleGAN, MUNIT~\citep{huang_multimodal_2018}, InstaGAN ~\citep{mo_instagan_2018}, InstaGAN using the mask to constrain the transformation to only the masked area (similarly to~\cref{eq:copy-paste}), and the ClimateGAN Painter applied to ground segmentation masks predicted by HRNet~\citep{tao2020ierarchical}. These models were trained using the original papers' configurations, with hyper-parameter tuning to be fair in our comparison. Some samples from the models can be seen in~\cref{fig:comparables}, and further results are provided in~\cref{sec:supp-human-eval}, ~\cref{fig:humeval-supp}.

\section{Results} 
\label{sec:results}

This section presents the model evaluation results, including the Masker ablation study and the comparison of the overall proposed model---Masker and Painter---against comparable approaches via human evaluation. Visual examples of the inferences of our model can be seen in ~\cref{fig:climategan-inferences}.

\subsection{Masker Evaluation}
\label{sec:masker-results}

\begin{figure}[bt]
\centering
\includegraphics[width=0.75\textwidth]{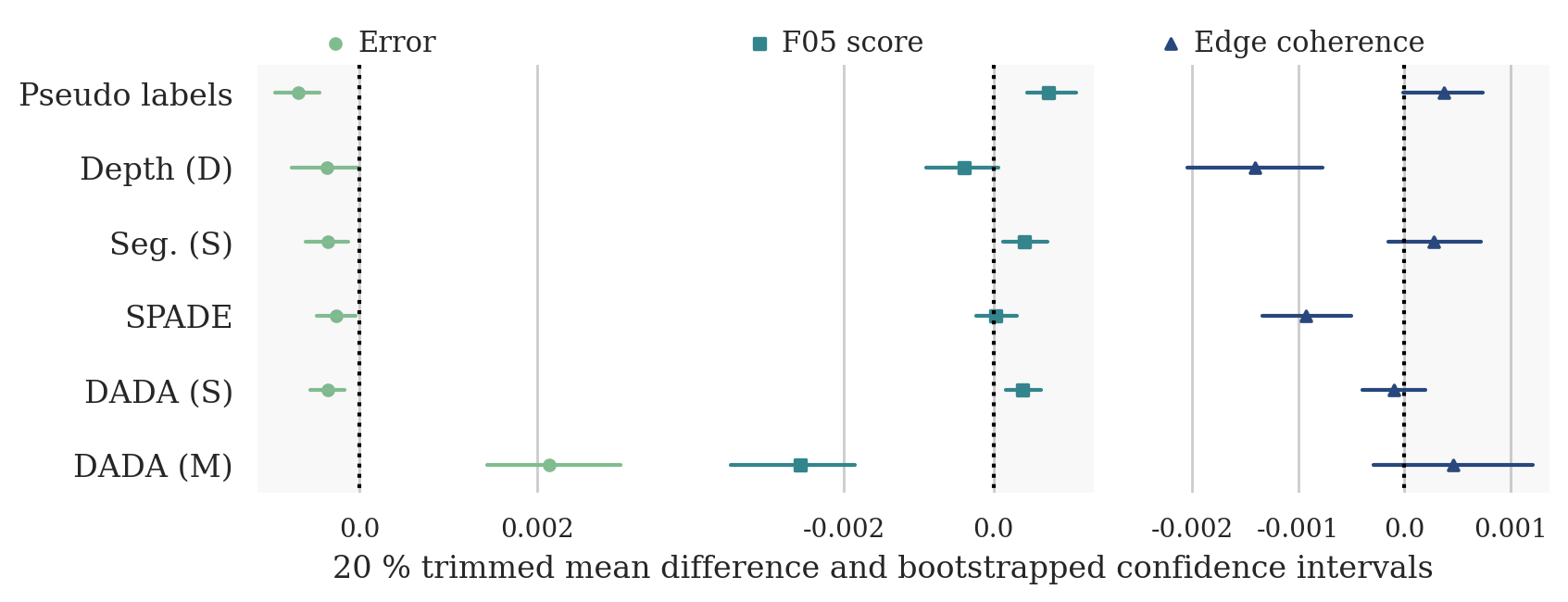}
\caption{Summary of the statistical inference tests of the ablation study. The shaded area indicates metric improvement. All techniques but DADA ($M$) significantly improved the error rate, and some further improved the F05 score and edge coherence.}
\label{fig:ablation-bootstrap-ci}
\end{figure}

The main conclusion of our ablation study is that five of the six techniques proposed to improve the quality of the Masker positively contribute to the performance. In~\cref{fig:ablation-bootstrap-ci}, we show the median differences and confidence intervals obtained through the bootstrap. The error rate improved---with 99~\% confidence---in the models that included pseudo labels (for the first 10 training epochs), a depth and a segmentation head, SPADE and DADA for the segmentation head---recall that the contribution of each technique was tested separately. For some but not all techniques, the F05 score and edge coherence also improved significantly. In contrast, we found that both the error and the F05 score were worse when DADA for the Masker was included.

\begin{figure}[th]
\centering
\includegraphics[width=\textwidth]{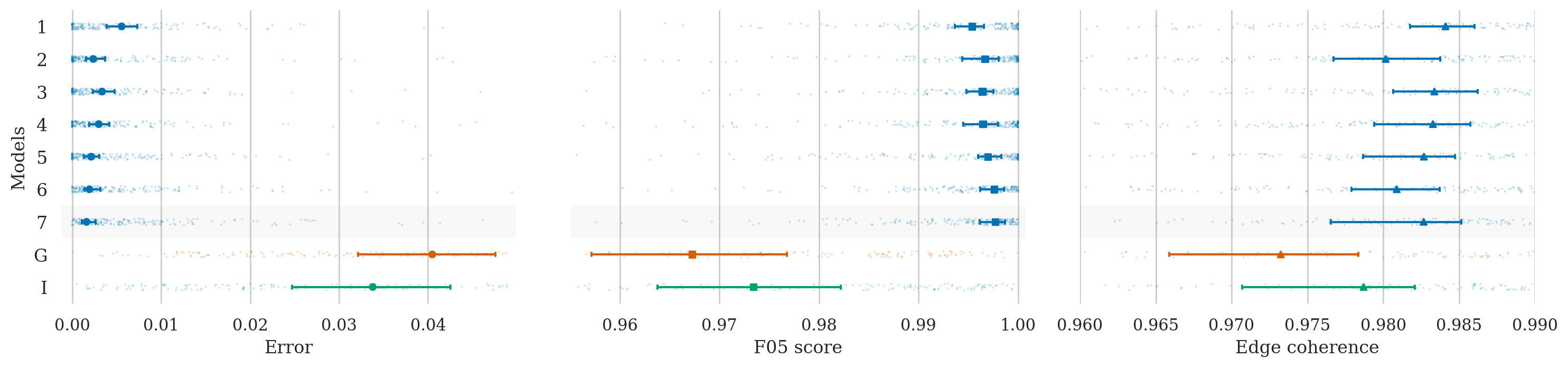}
\caption{Evaluation metrics for a subset of the models studied in the ablation study and presented in~\cref{tab:ablation}---models trained without pseudo labels or with DADA for the Masker are excluded---as well as the two baselines for comparison. The solid symbols indicate the median of the distribution and the error lines the bootstrapped 99~\% confidence intervals. The shaded area highlights the best model---7.}
\label{fig:ablation}
\end{figure}

In \cref{fig:ablation}, we show the Masker evaluation metrics for a subset of the models, 1--7, selected upon the conclusions of the ablation study, as well as the two baseline models. The main conclusion of the evaluation is that our proposed Masker largely outperforms the baselines for the three metrics, especially in terms of the error and the F05 score. Also, the metrics of the individual models support the findings of the ablation study, as the best performance---5, 6 and 7---is achieved by the models that include all or most techniques: pseudo labels, depth and segmentation heads, DADA for $S$ and SPADE. Surprisingly, model 2, which only includes pseudo labels and the depth head, also achieves high performance. In contrast, models 3 and 4, which include a segmentation head without DADA or SPADE, obtain worse performance. Therefore, it seems that the contribution of the segmentation head is clearly complemented by using DADA and SPADE. We include a complete analysis of the ablation study and Masker evaluation metrics in \cref{sec:suppl:masker-evaluation}. The final architecture we selected for the Masker, 7, includes: pseudo labels, $D$, $S$, DADA for $S$ and SPADE for $M$.

\subsection{Human Evaluation}
\label{sec:humaneval}

\begin{figure}[hbt]
\centering
\includegraphics[width=0.75\textwidth]{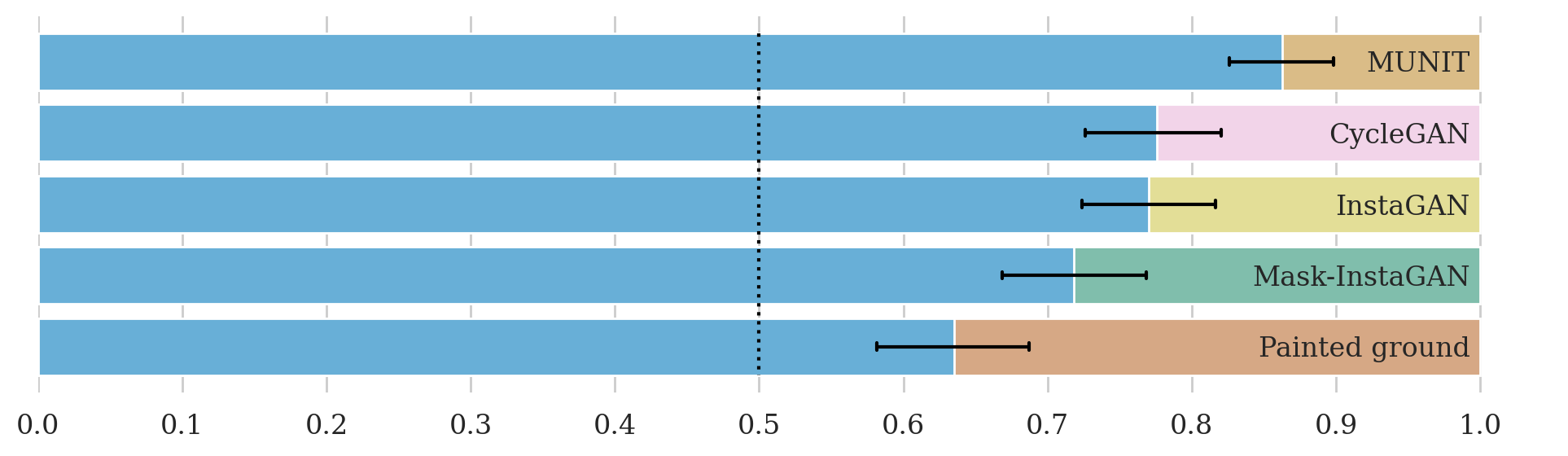}
\caption{Results of the human evaluation: the blue bars indicate the rate of selection of ClimateGAN over each alternative. The error lines indicate 99~\% confidence intervals.}
\label{fig:humeval}
\end{figure}

The difficulty of evaluating the performance of generative models is a well-known challenge in the field~\citep{theis2015note, zhou2019hype}. Commonly used metrics for evaluating the quality of generated images, such as the Fréchet Inception Distance (FID)~\citep{heusel2017ans}, need a large number of samples to be statistically significant~\citep{Seitzer2020FID}, which is not feasible in our low data regime. Therefore, in order to compare ClimateGAN to related approaches, we carried out a survey with human evaluators using the \href{https://www.mturk.com/}{Amazon Mechanical Turk} platform. We presented evaluators with pairs of images generated based on the same input image---one created by ClimateGAN, the other by one of the five related works described in \cref{fig:comparables}. We asked them to pick which image looks more like an actual flood, collecting 3 evaluations per pair. We present the results of this procedure in~\cref{fig:humeval}, where we can observe that ClimateGAN is consistently preferred by a significant margin to all other models. Interestingly, the single fact of constraining transformations to the masked area improves the performance of InstaGAN, and we hypothesize that this operation contributes to the performance of ClimateGAN. In addition, we can see that the second-best model is the Painted Ground, meaning that even with an off-the-shelf ground segmentation model to produce masks, the Painter is good enough on its own to produce realistic images, but that it performs best when paired with the Masker.

\section{Limitations and Improvements}
\label{sec:limitations-and-improvements}

An intuitive extension of our model would be to render floods at any chosen \textit{height}. Despite the appeal of this approach, its development is compromised by data challenges. To our knowledge, there is no data set of metric height maps of street scenes, which would be necessary for converting \emph{relative} depth and height maps into \emph{absolute} ones. Moreover, simulated worlds---including our own---that have metric height maps do not cover a large enough range of scenes to train models that would generalize well on worldwide Google Street View images. Another promising direction for improvement would be to achieve multi-level flooding, controlling water level represented by a mask, which faces the same challenges.

We also further explored the integration of multi-task learning in our approach, attempting to apply several of the methods mentioned in \citep{crawshaw2020multitask} to weigh the various Masker losses, including dynamic weight average~\citep{Liu_2019_CVPR} and weighting losses by uncertainty~\citep{Kendall_2018_CVPR}. We found that our manual tuning of constant weights performs better than the aforementioned methods. In future work, we aim to try more advanced techniques like gradient modulation methods \citep{NEURIPS2020_3fe78a8a, 8954118} to explore the further integration of multi-task learning to our approach.

\section{Conclusion}
\label{sec:conclusion}

In the present article, we have proposed to leverage advances in modern generative modeling techniques to create visualizations of floods, in order to raise awareness about climate change and its dire consequences. In fact, the approach described is part of a wider initiative aiming to create an interactive web-based tool that allows users to input the address of their choice and visualize the consequences of extreme climate events like floods, while learning about the causes and impacts of climate change.

Our contributions comprise a data set of images and labels from a 3D virtual world, an architecture we call ClimateGAN, and a thorough evaluation method to measure the performance of our system. We found our proposed Masker-Painter dichotomy to be superior to existing comparable techniques of conditional generations in two ways: first, we empirically showed that we are able to produce more realistic flood masks by informing the flood mask decoder with geometrical information from depth predictions and semantic information from segmentation maps, and by training the three decoders together. Second, we established, using human evaluations, that our Painter alone is able to create realistic water textures when provided ground masks, and that its performance increases even further when provided with the Masker's predictions. Overall, this allows ClimateGAN to produce compelling and robust visualizations of floods on a diverse set of first-person views of urban, suburban and rural scenes.

\clearpage
\bibliographystyle{bibliography}
\bibliography{references, references2}

%
%
% ------------------------- APPENDIX -------------------------
%
%

\clearpage
\appendix

{

\section{Simulated data}\label{sup:sim}

In this section, we expand on \cref{sec:simdata} and provide details about our simulated world and the images and labels we can obtain from it. 

We created a 1.5 $km^2$ virtual world using the Unity3D game engine containing urban, suburban and rural areas.

The $urban$ environment contains skyscrapers, large buildings, and roads, as well as objects such as traffic items and vehicles. \cref{fig:birdeye} shows a bird's eye view of the urban area of our virtual environment. The $rural$ environment consists of a landscape of grassy hills , forests, and mountains, with sparse houses and other buildings such as a church, and no roads. The rural and urban areas make up for 1 $km^2$ of our virtual world.

\begin{figure}[ht]
\begin{center}
\includegraphics[width=0.65\textwidth]{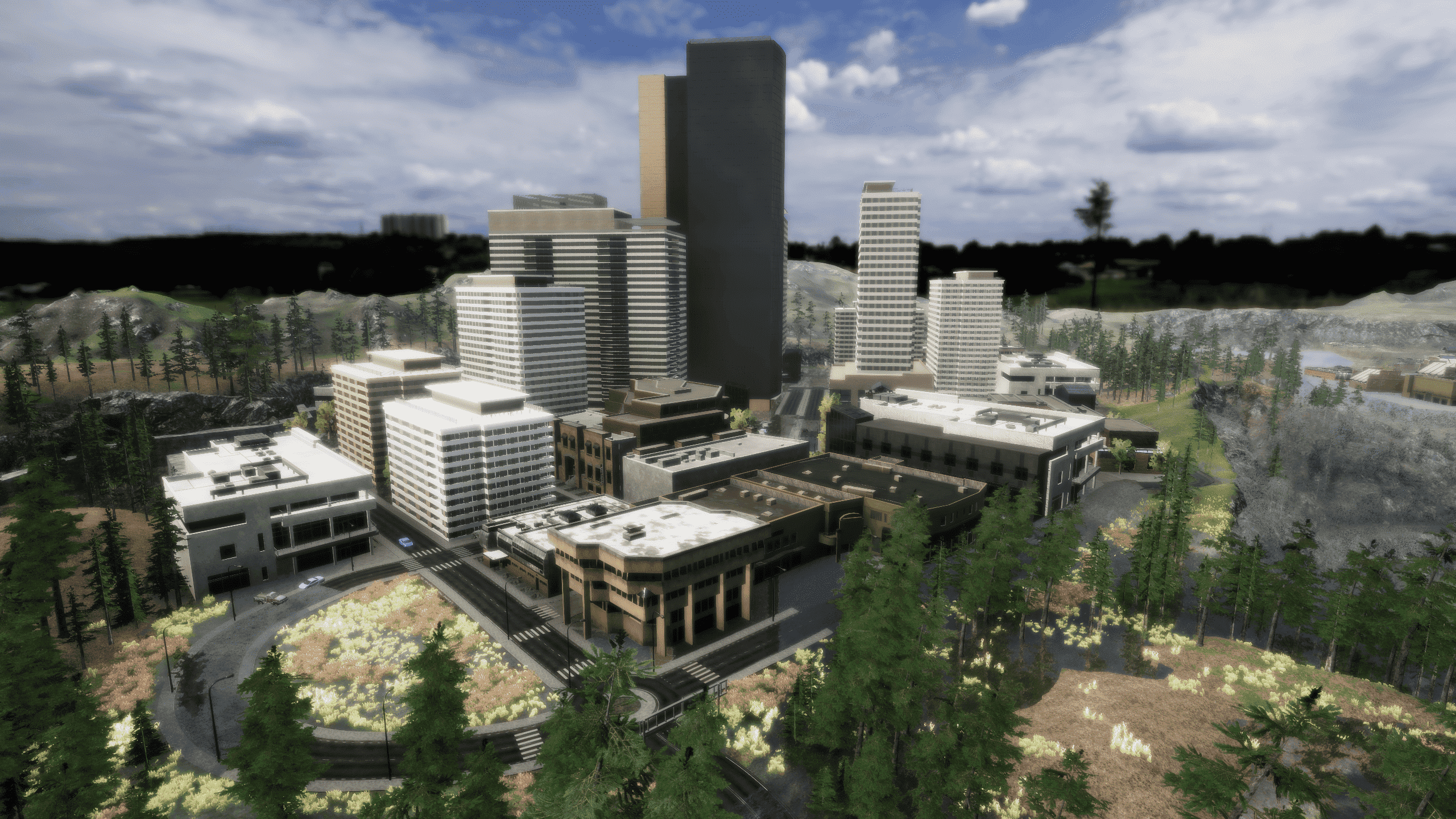}
\caption{Bird's eye view of the urban area (city) and rural area (outskirts of the city) of our simulated world}
\label{fig:birdeye}
\end{center}
\end{figure}

The $suburban$ environment (Figure~\ref{fig:birdeye_sub}) is a residential area of 0.5 $km^2$ with many individual houses with front yards. 

\begin{figure}[ht]
    \centering
    \begin{subfigure}[b]{0.48\linewidth}
      \centering
      \includegraphics[width=\textwidth]{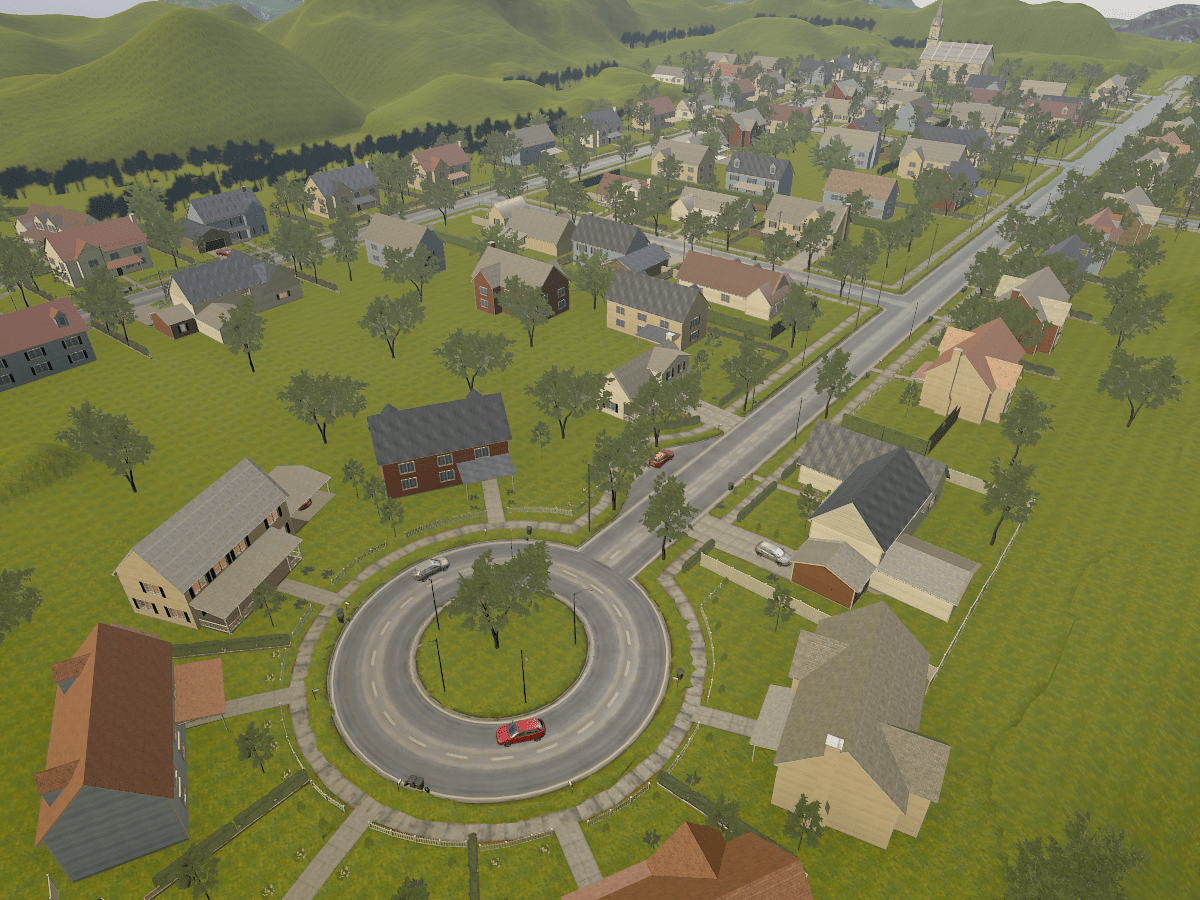}  
    \end{subfigure}
    \hfill
    \begin{subfigure}[b]{0.48\textwidth}
      \centering
      \includegraphics[width=\textwidth]{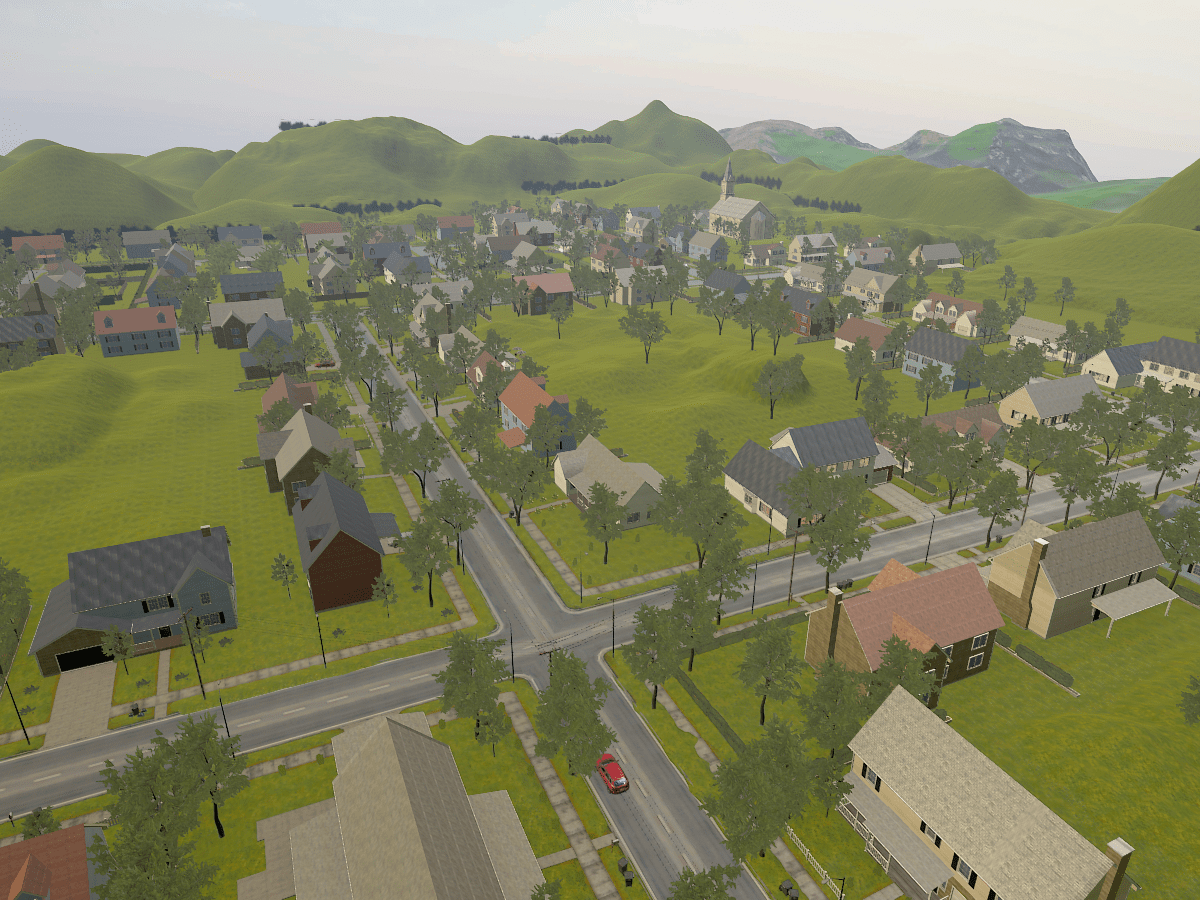}  
    \end{subfigure}
\caption{Bird's eye views of the suburban area of our simulated world}
\label{fig:birdeye_sub}
\end{figure}

To gather the simulated dataset, we captured `before' and `after' flood pairs from 2000 viewpoints with the following modalities: 
\begin{itemize}
\setlength{\itemsep}{1pt}
  \setlength{\parskip}{0pt}
  \setlength{\parsep}{0pt}
  \item `before' : non-flooded RGB image, depth map, segmentation map
  \item `after' : flooded RGB image, binary mask of the flooded area, segmentation map
 \end{itemize}
The camera was placed about $1.5m$ above ground, and has a field of view of 120$^\circ$, and the resolution of the images is 1200$\times$900. At each viewpoint, we took 10 pictures, by varying slightly the position of the camera in order to augment the dataset. 

\cref{fig:sim} shows the different modalities captured at each viewpoint, and \cref{fig:samples} shows samples of our simulated dataset in urban, suburban and rural areas.

\paragraph{Depth} The depth maps are provided as RGB images for the `before' case, and the depth is recorded up to $1000 m$ away from the camera, with precision of $4 mm$. 

\paragraph{Segmentation}
There are nine different classes of objects in the simulated world:
\begin{itemize}
\setlength{\itemsep}{1pt}
  \setlength{\parskip}{0pt}
  \setlength{\parsep}{0pt}
    \item \textit{sky}
    \item \textit{ground}: road, sidewalks, road markings, anything that is asphalt
    \item \textit{building}
    \item \textit{traffic item}: lampposts, traffic signs, poles 
    \item \textit{vegetation}: small bushes, trees, hedges excludes grass, lawns 
    \item \textit{terrain}: rocks, soil, lawns
    \item \textit{car}: cars and trucks
    \item \textit{other}: miscellaneous objects such as postboxes, trashcans, garbage bags, etc.
    \item \textit{water}: only present in the `after' flooded images
\end{itemize}

%We did not want to put our own biases into creating people
While people are not included in the simulated world, the segmentation model is able to learn this class from the real world due to the supervision signal given by the HRNet pseudo-labels.

\paragraph{Mask} We also include binary masks of the flood (water segmentation) for the `after' images. The masks are used to train the Masker with ground truth target flood masks in the simulated domain. 

\begin{figure}[ht]
\centering
\small
\begin{tabularx}{\textwidth}{*{4}{Y}}
RGB & Segmentation & Mask & Depth \\
\end{tabularx}
\includegraphics[width=\textwidth]{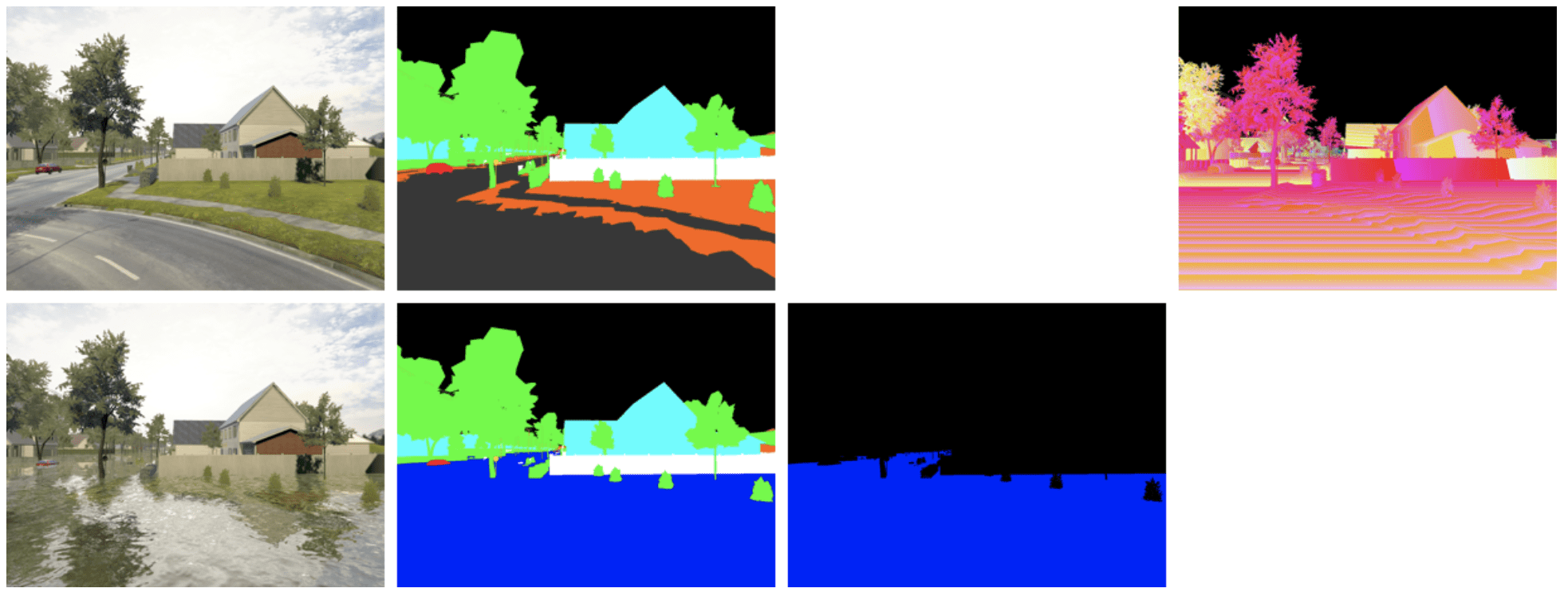}
\caption{Sample data obtained at one spot for one camera position in our virtual world. The top row shows the modalities of the `before' flood image: RGB image of the scene, depth map and segmentation map; and the bottom row shows those obtained in the `after' configuration: RGB image, segmentation map and binary flood mask.}
\label{fig:sim}
\end{figure}   

\begin{figure}[h]
\centering
\small
% include first image
\includegraphics[width=\textwidth]{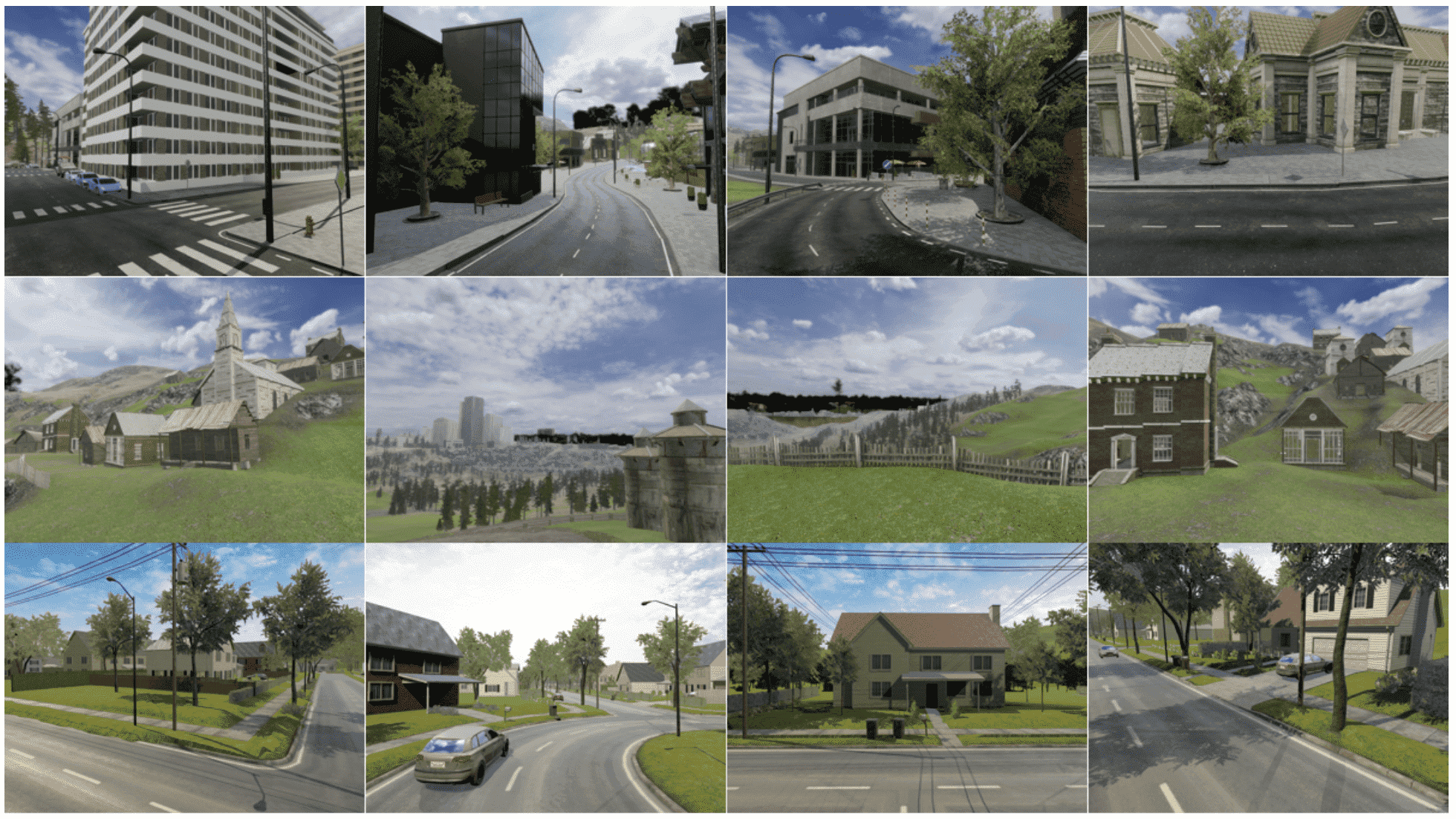}
\caption{Samples from our simulated dataset in urban (top row), rural (middle row), and suburban (bottom row) areas.}
\label{fig:samples}
\end{figure}

% \pagebreak

\section{Losses}
\label{sup:losses}

For clarity, in \cref{sec:masker} we did not detail the exact formulation of some of the losses since they are either straight-forward (as is the Cross Entropy) or direct applications of their definitions as per their original paper (\textit{e.g.} the SSIMSE loss). We expand on those here.

\subsection{Depth Decoder}
In this section, we detail the depth decoder loss presented in \cref{sec:masker} and adapted from~\citep{midas}.\\
Let $d$ a predicted disparity map and $d^*$ the corresponding ground truth. 
The aligned disparity maps with zero translation and unit scale are: 
\begin{equation}
 \hat{d} = \frac{d-\textbf{t}(d)}{\textbf{s}(d)} \text{  ,  }\hat{d}^* = \frac{d^*-\textbf{t}(d^*)}{\textbf{s}(d^*)}
 \end{equation}
where \textbf{t} and \textbf{s} are defined as $\textbf{t}(d) = \text{median}(d)$ and $\textbf{s}(d) = \frac{1}{N}\sum_{n}|d - \textbf{t}(d)|$.\\

The depth decoder loss is composed of a scale-and-shift invariant MSE loss term : 
\begin{equation}
\begin{split}
\mathcal{L}_{SSIMSE} &= \frac{1}{2} \E_{n,h,w}[(\hat{d}^{(n,h,w)} - \hat{d}^{*(n,h,w)})^2]
\end{split}
\end{equation}

an a multi-scale gradient matching term to enforce smooth gradients and sharp discontinuities: 
\begin{equation}
\begin{split}
\mathcal{L}_{GM} &=   \E_{n}[\sum_{k}\sum_{h,w}|\nabla_x R_{k}^{(n,h,w)}|+|\nabla_y R_{k}^{(n,h,w)}|]
\end{split}
\end{equation}

where $R^{(n,h,w)} =\hat{d}^{(n,h,w)} - \hat{d}^{*(n,h,w)}$ and $R_{k}$ corresponds to the difference of disparity (inverse depth) maps at scale $k$. \\
Following ~\citep{midas}, we consider 4 scale levels, downsampling by a factor two the image at each scale. 

\subsection{Segmentation Decoder}

\comment{For an input $\vx_i$, we start by computing the segmentation map $\vs_i = S(E(\vs_i),\vd_i)$ and the depth-aware map $\hat{I}_{s_i}$,} 

This decoder computes the segmentation map $\vs_i$ given an input image $\vx_i$, $\vs_i$ being a 4D tensor of shape $N \times C \times H \times W$. The number of channels $C$ corresponds to the number of classes, nine in our case: \textit{ground}, \textit{building}, \textit{traffic item}, \textit{vegetation}, \textit{terrain}, \textit{car}, \textit{sky}, \textit{person} and \textit{other}.

\comment{The segmentation decoder $S$ is based on DeepLabv3+ architecture \citep{deeplabv3plus2018}. Also, as proposed in DADA paper paper~\citep{vu_dada_2019}, we leverage depth information available in the simulated world to improve our segmentation predictions by giving more attention to closer objects.} First, we detail the two fusion mechanisms used in the DADA approach \citep{vu_dada_2019}: 
\begin{itemize}
    \item \textbf{Feature fusion}: It is the element-wise multiplication between the latent vector and a depth vector of the same size. This depth vector is obtained by a 1 x 1 convolutional layer applied to the depth head before the average pooling. The fused features are given as input to the segmentation decoder.
    \item \textbf{DADA fusion}: Instead of giving the self-information map $I^S_i$ to the AdvEnt~\citep{vu_advent_2018} discriminator $Q^S$, we give it the element-wise multiplication between $I^S_i$ and the depth predictions $\vd_i$. The obtained matrix is called the depth-aware map  $\hat{I}^S_i$.
\end{itemize} 

We also detail all segmentation losses that are not defined in the main paper. First, the cross-entropy loss is defined as:

\comment{
\begin{equation}
\begin{split}
\mathcal{L}_{CE}(\vy_i, \vs_i) &= -\E_{n,h,w}[ \sum_{c} \Bigl(\vy_i^{(n, c, h, w)} \log \vs_i^{(n, c, h, w)} \Bigr)]
\end{split}
\end{equation}
}

\begin{equation}
\begin{split}
\mathcal{L}_{CE}(\vy_i, \vs_i) &= -\sum_{c}\E_{n,h,w}[  \vy_i^{(n, c, h, w)} \log \vs_i^{(n, c, h, w)}]
\end{split}
\end{equation}

\comment{
\begin{equation}
\begin{split}
\mathcal{L}_{CE}(\vy_i, \vs_i) &= -\E_{n,c, h,w}[\vy_i^{(n, c, h, w)} \log \vs_i^{(n, c, h, w)}]
\end{split}
\end{equation}
}

Furthermore, the entropy minimization loss, which is used as suggested by the authors of ADVENT\citep{vu_advent_2018}, is computed according to the following equation:

\comment{
\begin{equation}\label{eq:ME}
\begin{split}
\mathcal{L}_{EM} (\vs_r)= \frac{-1}{\log(C)} \E_{n,h,w}[\sum_{c} \vs_r^{(n, c, h, w)}\log \vs_r^{(n,c, h, w)}]
\end{split}
\end{equation}
}

\comment{
\begin{equation}\label{eq:ME}
\begin{split}
\mathcal{L}_{EM} (\vs_r)= \frac{-1}{\log(C)}\sum_{c}  \E_{n,h,w}[\vs_r^{(n, c, h, w)}\log \vs_r^{(n,c, h, w)}]
\end{split}
\end{equation}
}

\begin{equation}\label{eq:ME}
\begin{split}
\mathcal{L}_{EM} (\vs_r)= \E_{n,c,h,w}[-\vs_r\log \vs_r]
\end{split}
\end{equation}

Finally, we detail the WGAN loss that is computed using a discriminator $Q_s$ which outputs the probability of the depth-aware map coming from the real domain. If updating the decoder $S$, the loss tries to fool the discriminator:

\begin{equation}\label{eq:G-GAN-S}
    \mathcal{L}_{G-GAN}(Q^S, \hat{I}(\vs_r), \hat{I}(\vs_s)) = - \E_{n}[ Q^S(\hat{I}(\vs_r))]
\end{equation}

If updating D, the loss tries to improve the discriminator's prediction:

\begin{equation}\label{eq:D-GAN}
    \mathcal{L}_{D-GAN} (Q^S, \hat{I}(\vs_r), \hat{I}(\vs_s)) = -\E_{n} [Q^S(\hat{I}(\vs_s)) - Q^S(\hat{I}(\vs_r))]
\end{equation}

\subsection{Flood Mask Decoder}
Similarly to the segmentation decoder, we detail in this section the WGAN loss that is computed using a discriminator $Q^M$ in order to train M. This discriminator outputs the probability of the self-information map $I(\vm_i)$ coming from the real domain. When updating the decoder $M$, the loss tries to fool the discriminator:

\begin{equation}\label{eq:G-GAN-M}
\begin{aligned}
   \mathcal{L}_{G-GAN}(Q^M, I(\vm_r), I(\vm_s)) = - \E_{n}[ Q^M(I(\vm_r))]
\end{aligned}
\end{equation}

When updating $Q^M$, the loss tries to improve the discriminator's predictions:

\begin{equation}
  \begin{aligned}
    \mathcal{L}_{D-GAN} & (Q^M, I(\vm_r), I(\vm_s)) = \\
    & -\E_{n} [Q^M(I(\vm_s)) - Q^M(I(\vm_r))]
  \end{aligned}
\end{equation}

\comment{
The complete loss is then:

\begin{equation}\label{eq:GAN}
    \mathcal{L}_{GAN}(Q_m, I_{m_r}, I_{m_s}) = \begin{cases}
    \mathcal{L}_{G-GAN}(Q_m, I_{m_r}, I_{m_s}) \text{ if updating }G \\
    \mathcal{L}_{D-GAN}(Q_m, I_{m_r}, I_{m_s}) \text{ if updating }Q_m \\
    \end{cases}
\end{equation}
}

\section{Masker Evaluation}
\label{sec:suppl:masker-evaluation}

In \cref{sec:masker-evaluation}, we introduced the method for evaluating the quality of Masker's output. Here, we provide details of the annotation procedure for generating a Masker test set (\cref{sup:test-annotations}) and the metrics proposed to assess the performance (\cref{sec:suppl:metrics}).

\subsection{Test Set Collection and Annotation}\label{sup:test-annotations}

In this section, we provide more details about our test set data collection procedure, and the annotation guidelines we followed. \\
We collected the 180 images of our test set  from Google Street View with the following selection guidelines:

\begin{itemize}
\setlength{\itemsep}{1pt}
  \setlength{\parskip}{0pt}
  \setlength{\parsep}{0pt}
    \item Geographical diversity: we selected images in a variety of cities on all continents, with different architectural styles and general landscapes. 
    \item Varied levels of urban development: both urban and rural areas were included.
    \item Variety of scenes content: we endeavored to cover a wide range of objects in our test set, including images containing bike racks, dense crowds, and various vehicles in urban areas, and different vegetation types in rural areas. 
\end{itemize}

We also carefully collected challenging images to determine the limitations of our model, including images facing slopes (going up or down), or stairs, with various ground textures, around areas under construction, and in areas near canals among others. 
The images were manually annotated, and the pixels of each image were categorized in one of three classes according to the following instructions: 

\begin{itemize}
\setlength{\itemsep}{1pt}
  \setlength{\parskip}{0pt}
  \setlength{\parsep}{0pt}
\item \emph{Must be flooded}: This class contains the minimal region that should be flooded. We want to represent floods of height at least 0.5 m. Typically, this corresponds to flooding up to the knees of adult pedestrians or up to the top of cars' wheels. In cases when no such reference objects were available, the annotator would make the 0.5m estimate based on other cues, such as doorsteps, traffic signs and vegetation.
\item \emph{Cannot be flooded}: This label indicates regions we absolutely do not want to be flooded. We generally put any pixel corresponding to an object with a height greater than ~1.5 m above ground in this category. This includes car roofs and adult pedestrians' heads, which we used as reference to determine the lower limit of this region.
\item \emph{May be flooded}: This category contains any pixel not assigned to the other classes. It reflects the fact that we do not enforce flooding at a specific height yet value plausible flood location. 
\end{itemize}

We show examples of the labeled test images in \cref{fig:labels}.

\begin{figure}[htb]
\begin{center}
\includegraphics[width=0.7\textwidth]{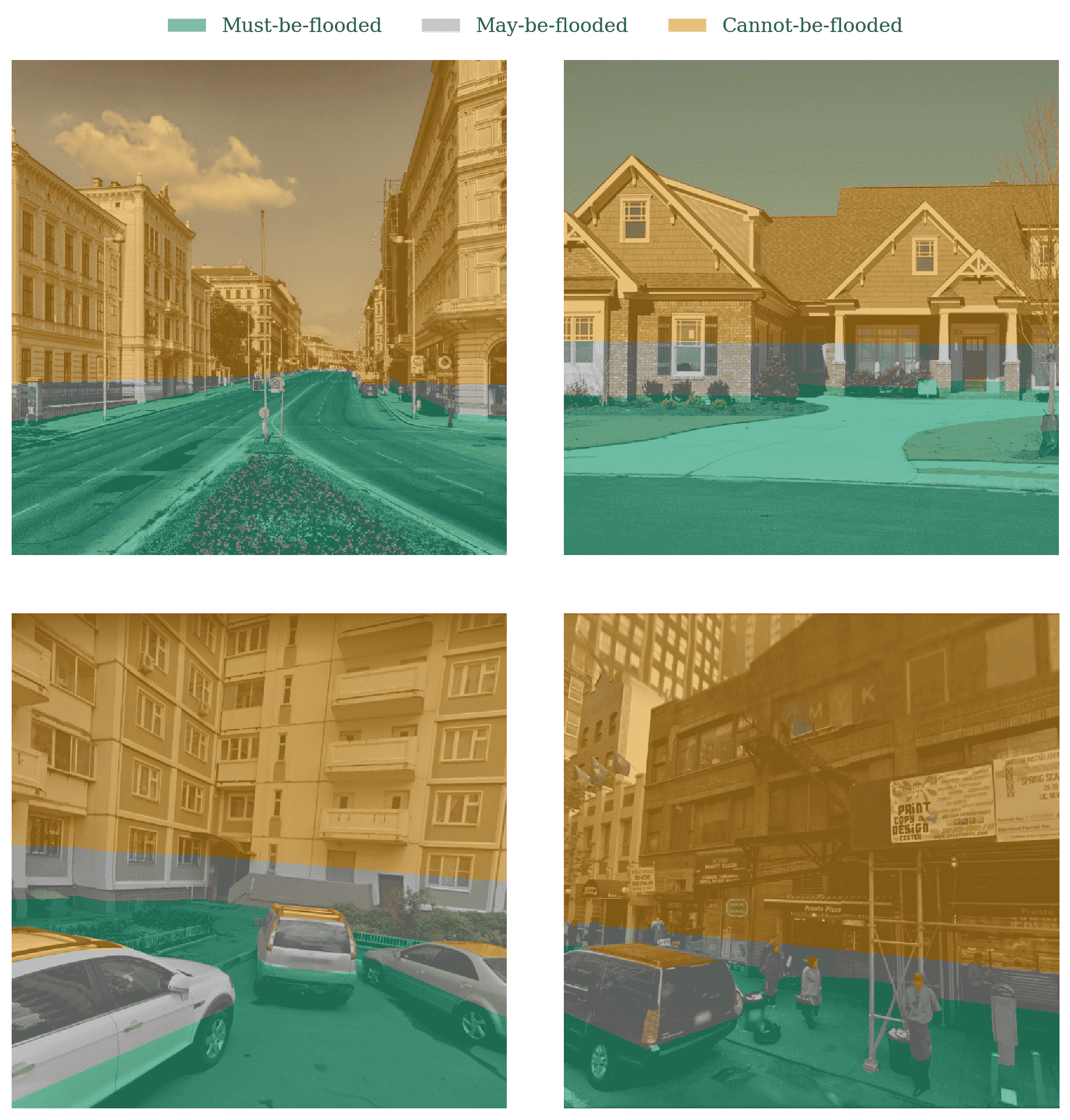}
\caption{Examples of labeled images from our test set.}
\label{fig:labels}
\end{center}
\end{figure}

\subsection{Metrics}
\label{sec:suppl:metrics}

In \cref{sec:metrics} we proposed three metrics to compare the masks predicted by the Masker and the labeled test images: error (\cref{eq:error}), F05 score and edge coherence (\cref{eq:edge-coherence}). Here, we delve into the reasons why we proposed more than one metric and the contribution of each metric.

We proposed the error---number of erroneously predicted pixels divided by the size of the image---as the main evaluation metric, since it characterizes the size of the errors in the image, which directly impacts the perceived mask quality.

However, we argue that the error alone does not capture all aspects of the performance of a Masker model in the test set. For example, the error, as defined in \cref{eq:error} does not take into account the size of the labeled areas (see image ``D'' in \cref{fig:metrics-all}). While the size of the labels may have a smaller perceptual impact, in order to characterize the precision and sensitivity of the model, additional metrics are useful. We proposed the F05 score:
\begin{equation}
\label{eq:f05}
\begin{gathered}
F05 = \frac{1.25 \times precision \times recall}{0.25 \times precision + recall} = \frac{1.25 \times \frac{TP}{TP+FP} \times \frac{TP}{TP+FN}}{0.25 \times \frac{TP}{TP+FP} + \frac{TP}{TP+FN}}
\end{gathered}
\end{equation}
which computes the weighted harmonic mean of precision and recall---also known as sensitivity or true positive rate. We used $\beta=0.5$ in order to weigh precision more than recall, that is to penalize more false positives than false negatives. In our context, this translates into setting a higher penalty for flooding \textit{cannot-be-flooded} pixels---for instance heads of pedestrians, automobile roofs and high areas of the image, in general---than for missing areas that should be flooded. While both types of errors should be penalized, the former has a higher perceptual impact.

\begin{figure}[!ht]
\centering
\includegraphics[width=0.7\textwidth]{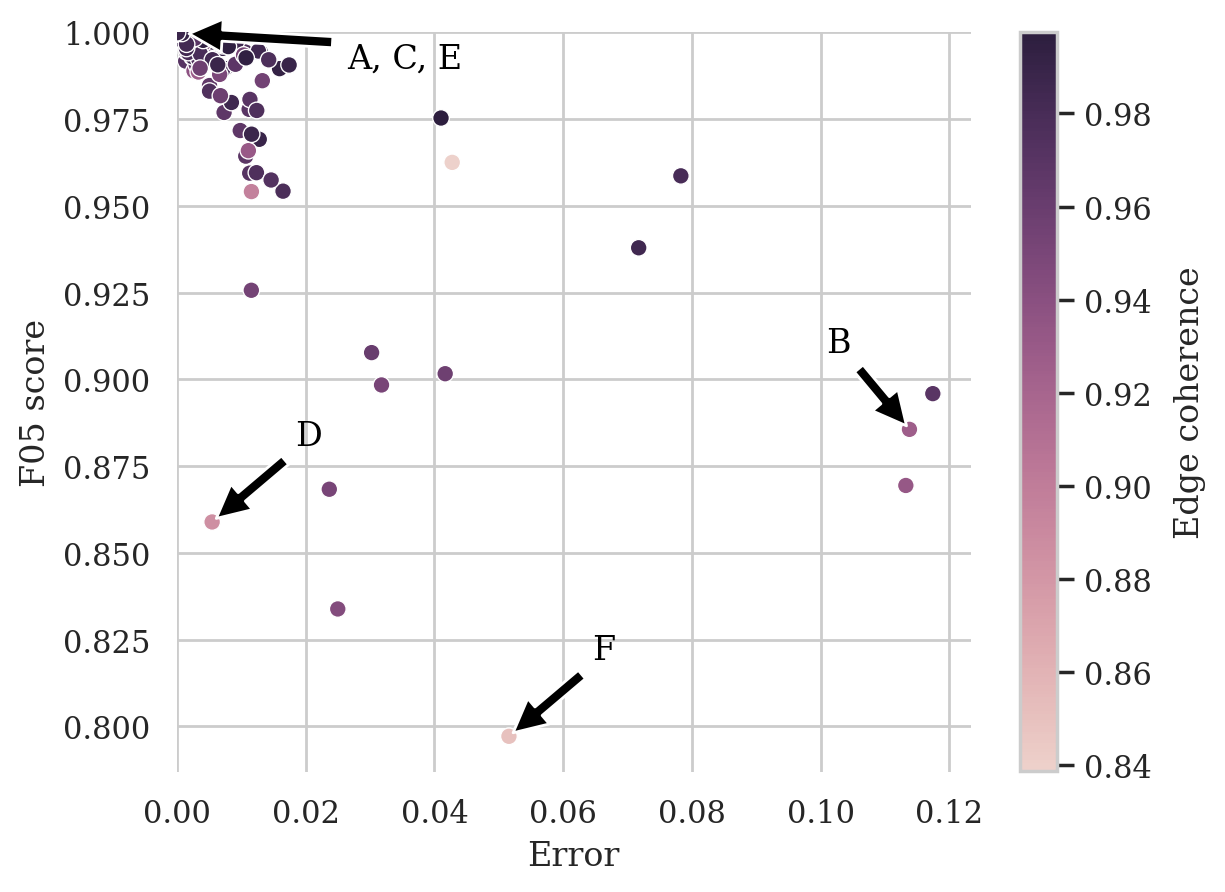}
\caption{Distribution of the three metrics for the selected best Masker. The annotations correspond to the images in \cref{fig:metrics-all}.}
\label{fig:scatter_metrics}
\end{figure}

Finally, we proposed an edge coherence metric (\cref{eq:edge-coherence}) in order to take into consideration the shape of the predicted mask, with respect to the shape of the \textit{must-be-flooded} label. Note that neither the error or the F05 score account for the shape, as only the amount of correct or incorrect pixels matter, regardless of the position. As explained in the previous section, we defined a \textit{may-be-flooded} class in order to allow for different levels of flooding in the mask prediction. That is, higher levels of flooding in a prediction should not be penalized necessarily, as long as the mask is consistent with the semantics in the image. Our proposed metric of edge coherence is based on the assumption that predictions whose border is roughly parallel to the border of the \textit{must-be-flooded} label should be less penalized than highly dissimilar shapes.

\cref{fig:scatter_metrics} shows the value of the three metrics for all the images in the test set. While there is certain correlation between the metrics, especially because the bulk of the distributions is around images for which the predictions are very accurate and hence all the metrics are near perfect, the plot shows that various metrics are useful for identifying a few images for which not all metrics are low. We further illustrate the meaning of our proposed metrics in \cref{fig:metrics-all}, where we show images that obtained the lowest---2nd quantile---and highest---98th quantile---values of each metric.

\begin{table}[]
\centering
\begin{tabular}{@{}lll@{}}
\toprule
              & Error              & p                \\ \midrule
Pseudo labels & $-6.9\times 10^{-4}$ {{[}}$-9.5\times 10^{-4}$, $-4.5\times 10^{-4}${{]}}   & = 0.0            \\
Depth         & $-3.7\times 10^{-4}$ {{[}}$-7.3\times 10^{-4}$, $-1.4\times 10^{-5}${{]}}   & \textless 0.01   \\
Seg. (S)      & $-3.6\times 10^{-4}$ {{[}}$-6.0\times 10^{-4}$, $-1.2\times 10^{-4}${{]}}   & \textless 0.0001 \\
SPADE         & $-2.6\times 10^{-4}$ {{[}}$-4.7\times 10^{-4}$, $-5.1\times 10^{-5}${{]}}   & \textless 0.01   \\
DADA (S)      & $-3.5\times 10^{-4}$ {{[}}$-5.5\times 10^{-4}$, $-1.7\times 10^{-4}${{]}}   & = 0.0            \\
DADA (M)      & $+2.1\times 10^{-3}$ {{[}}$+1.5\times 10^{-3}$, $+2.9\times 10^{-3}${{]}}   & = 0.0            \\ \bottomrule \\
\end{tabular}
\caption{Details of the results of the ablation study. The values in each cell are the 20~\% trimmed mean \textit{error} difference between models with and without a given technique, in brackets the 99~\% confidence intervals, and the $p$ value.}
\label{tab:ablation-bootstrap-ci-pval}
\end{table}
 
\begin{table}[]
\centering
\begin{tabular}{@{}lll@{}}
\toprule
              & F05 score  & p                                             \\ \midrule
Pseudo labels & $+7.4\times 10^{-4}$ {{[}}$+4.3\times 10^{-4}$, $+1.1\times 10^{-3}${{]}} & = 0.0             \\
Depth         & $-3.9\times 10^{-4}$ {{[}}$-9.1\times 10^{-4}$, $+5.9\times 10^{-5}${{]}} & \textless 0.1     \\
Seg. (S)      & $+4.1\times 10^{-4}$ {{[}}$+1.2\times 10^{-4}$, $+7.3\times 10^{-4}${{]}} & \textless 0.001   \\
SPADE         & $+2.9\times 10^{-5}$ {{[}}$-2.6\times 10^{-4}$, $+2.9\times 10^{-4}${{]}} & \textgreater 0.01 \\
DADA (S)      & $+3.8\times 10^{-4}$ {{[}}$+1.3\times 10^{-4}$, $+6.5\times 10^{-4}${{]}} & \textless 0.0001  \\
DADA (M)      & $-2.6\times 10^{-3}$ {{[}}$-3.5\times 10^{-3}$, $-1.8\times 10^{-3}${{]}} & = 0.0             \\ \bottomrule \\
\end{tabular}
\caption{The analogue to \cref{tab:ablation-bootstrap-ci-pval} for the F05 score}
\label{tab:ablation-bootstrap-ci-pval-f05}
\end{table}

\begin{table}[]
\centering
\begin{tabular}{@{}lll@{}}
\toprule
              &  Edge coherence  & p                                                        \\ \midrule
Pseudo labels & $+3.7\times 10^{-4}$ {{[}}$+1.2\times 10^{-5}$, $+7.4\times 10^{-4}${{]}} &   \textless 0.01      \\
Depth         & $-1.4\times 10^{-3}$ {{[}}$-2.1\times 10^{-3}$, $-7.4\times 10^{-4}${{]}} &   = 0.0                           \\
Seg. (S)      & $+2.8\times 10^{-4}$ {{[}}$-1.7\times 10^{-4}$, $+7.3\times 10^{-4}${{]}} &   \textgreater 0.01  \\
SPADE         & $-9.3\times 10^{-4}$ {{[}}$-1.4\times 10^{-3}$, $-5.0\times 10^{-4}${{]}} &   = 0.0                           \\
DADA (S)      & $-1.0\times 10^{-4}$ {{[}}$-4.2\times 10^{-4}$, $+2.2\times 10^{-4}${{]}} &   \textgreater 0.01 \\
DADA (M)      & $+4.6\times 10^{-4}$ {{[}}$-2.8\times 10^{-4}$, $+1.2\times 10^{-3}${{]}} &   \textgreater 0.01  \\ \bottomrule \\
\end{tabular}
\caption{The analogue to \cref{tab:ablation-bootstrap-ci-pval} for the edge coherence}
\label{tab:ablation-bootstrap-ci-pval-ec}
\end{table}

\begin{figure}[!ht]
\centering
\includegraphics[width=\textwidth]{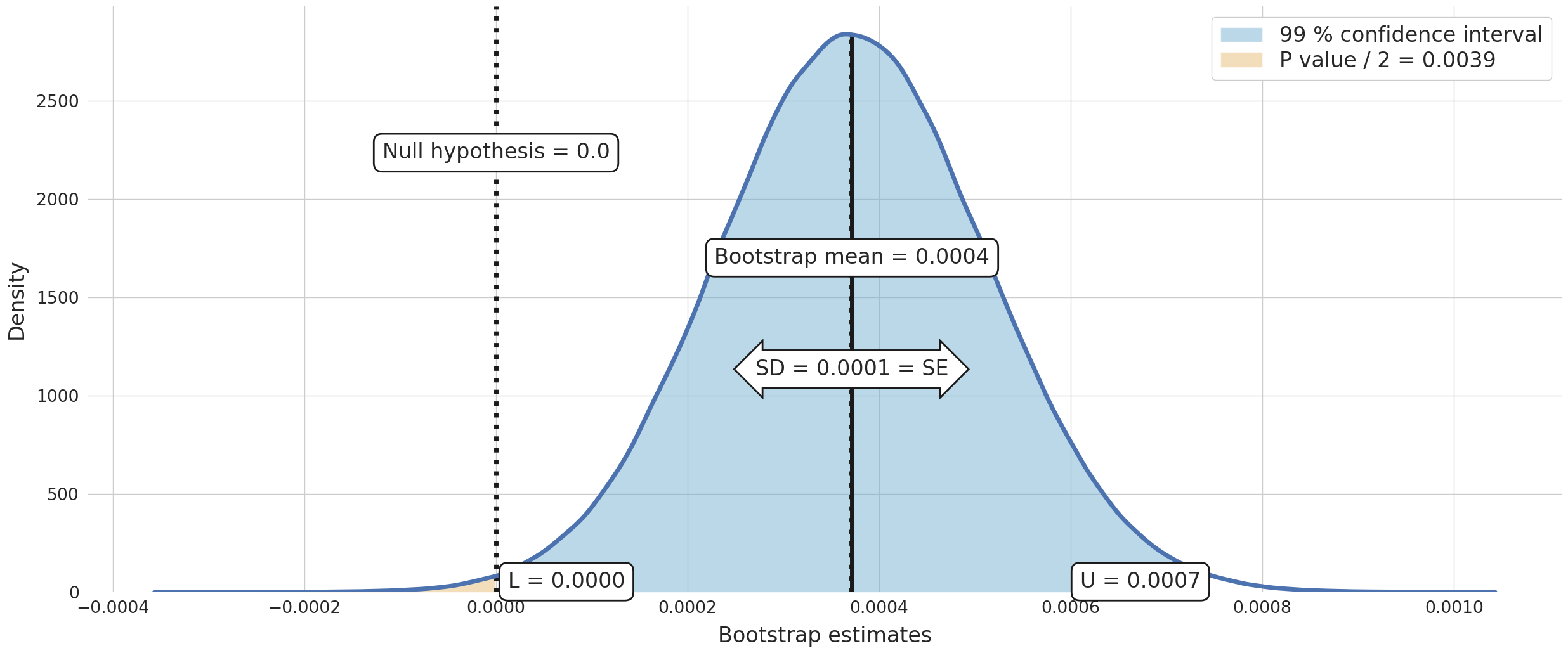}
\caption{Bootstrapped distribution of the 20~\% trimmed means of the difference in edge coherence between models that included pseudo labels and their counterparts. Equivalent distributions were obtained for all other techniques and metrics in the ablation study.}
\label{fig:bootstrap-distr}
\end{figure}

\subsection{Ablation Study}
\label{sec:suppl:ablation}

Here, we detail the methodology used for the ablation study of the Masker, presented in \cref{sec:ablation-method}, as well extend the set of results provided in \cref{sec:masker-results}.

In the ablation study, we studied the contribution to the Masker evaluation metrics of each technique: training with pseudo labels, a depth head ($D$), a segmentation head ($S$), SPADE, DADA for the segmentation head and DADA for the masker head. For every technique $t$, we considered all models $m_{j}^{t}$ which included such technique, and the paired models $m_{j}^{t_{0}}$ which differed only by the absence of $t$. Then, for every metric $r$, we constructed datasets of metric differences for every image $i$:
\begin{equation}
    d_{ij}^{r} = r(m_{j}^{t})_{i} - r(m_{j}^{t_{0}})_{i}
\end{equation}
\begin{figure}[!ht]
\centering
\includegraphics[width=\textwidth]{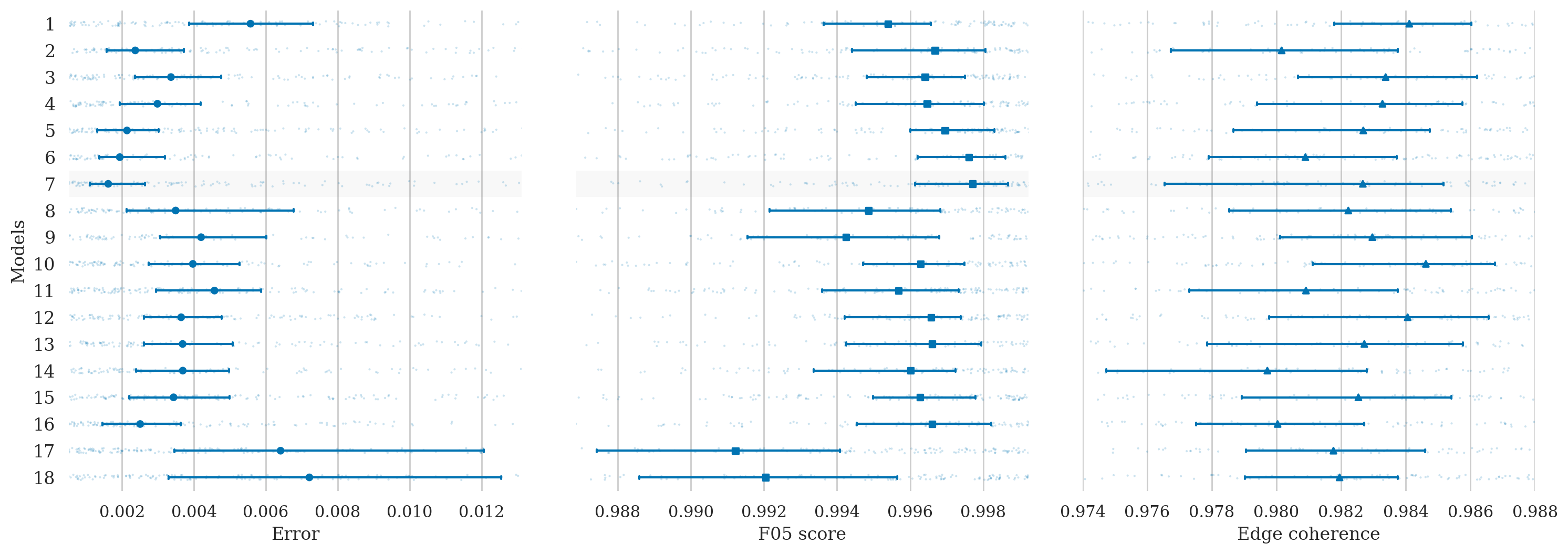}
\caption{Distribution of the metrics---error, F05 score and edge coherence---of the complete set of models tested in the ablation study, 1--18, excluding the two baselines, whose distributions are shown in \cref{fig:ablation}. The solid symbols indicate the median of the distribution and the error lines the bootstrapped 99~\% confidence intervals. The shaded area highlights the best model---7.}
\label{fig:ablation-all-no-baseline}
\end{figure}
and obtained 1 million bootstrapped samples. On each bootstrap sample we computed the 20~\% trimmed mean, which forms a bootstrap distribution from which we derived the confidence intervals we reported graphically in \cref{fig:ablation-bootstrap-ci}. In particular, we computed the 99~\% confidence intervals, that is the lower bound is the 0.5th quantile and the upper bound is the 99.5th quantile of the bootstrap distribution. In \cref{tab:ablation-bootstrap-ci-pval}, we provide the numerical details of all the tests in the ablation study. One advantage of the bootstrap over other statistical inference methods is that the outcome is a data-driven distribution rather than a binary test. In order to illustrate this, we provide the distribution for one of the techniques---pseudo labels---and one of the metrics---edge coherence---in \cref{fig:bootstrap-distr}.   

% \usepackage{booktabs}
% \usepackage{booktabs}
% \usepackage{booktabs}

In \cref{fig:ablation-all-no-baseline} we extend \cref{fig:ablation}, where we show the distribution of the metrics for the complete set of models in the ablation study. We do not include the baseline models, which obtained significantly worse metrics, in order to better visualize the differences within the models of the ablation study. From the figure, it becomes apparent that models trained without pseudo labels (9--18), and models trained with DADA for the masker head (8, 9, 17, 18) achieved worse performance, in general. This is consistent with the results of the ablation study obtained through the bootstrap.

\section{Human Evaluation Results}
\label{sec:supp-human-eval}

During the human evaluation of our results, we found that overall, evaluators primarily preferred ClimateGAN to other comparable approaches to varying degrees (a complete presentation of results can be found in~\cref{sec:humaneval}). However, there were a number of cases when evaluators preferred other approaches to ClimateGAN. We present such images in ~\cref{fig:humeval-supp}, where the images outlined in red are those that are those for which all 3 evaluators preferred a comparable approach to ClimateGAN.

It can be observed that it is hard to consider these approaches as being \emph{better} than ClimateGAN -- i.e., they do not appear to be more realistic neither in terms of the quality of generated water nor in the portion of the image that is flooded. However, the concept of realism and quality of generated imagery is, indeed, hard to define and to quantify, so it is unsurprising that some images were systematically preferred by evaluators. In order to carry out a more thorough human evaluation of our results, it would be necessary to have more evaluators per pair of images and, ideally, different experimental setups -- for instance, different captions and different designs of the comparison interface. Overall, it is hard to draw a conclusion for the motivation behind the preferences of individual evaluators. It can nonetheless be observed that in the case of some models, such as InstaGAN, images of light flooding (i.e. that after moderate rain) were preferred to that of more severe flooding generated by ClimateGAN. In other cases, such as that of InstaGAN+Mask, images with more pronounced reflections were chosen over those of the murkier, rippled water produced by ClimateGAN.

Finally, we feel that the best evaluation of the \emph{impact} of images generated by ClimateGAN is to compare them with other mediums of climate communication. For this purpose, we are working with a group of researchers in psychology to test the effect that our images have on climate risk perception. The experimental setup involves providing subjects with a text regarding the risks of climate change-induced flooding. One group of subjects will only see the text, a second will see the text accompanied by an image of a generic flooded house, whereas the third will see an image of their address of residence flooded using ClimateGAN. Preliminary results have indicated that individuals who saw their own place of residence flooded are more likely to perceive climate change as a real threat to their livelihood, and more likely to take action to fight it.

\begin{figure}
\begin{tabularx}{\textwidth}{*{7}{Y}}
\small{Input} & \small{CycleGAN} & \small{MUNIT} & \small{InstaGAN} & \small{InstaGAN+Mask} & \small{Painter+Ground} & \small{ClimateGAN}
\end{tabularx}
    \centering
    \includegraphics[width=\textwidth]{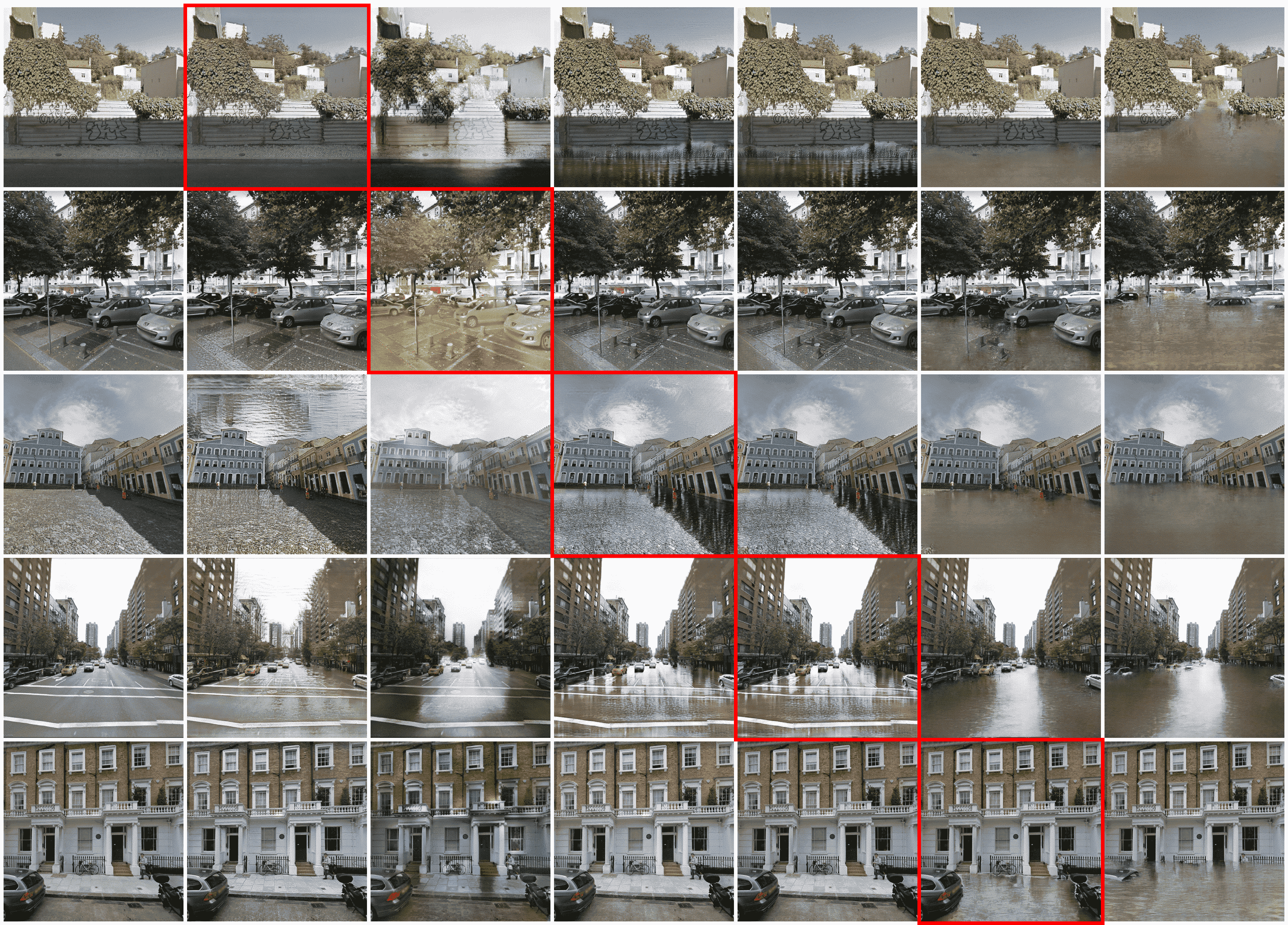}
    \caption{Example inferences of ClimateGAN and comparable approaches on images from the test set. Images outlined in red are those that were systematically (i.e. 3 out of 3 times) preferred over ClimateGAN.}
    \label{fig:humeval-supp}
\end{figure}

\section{Carbon Impact}

The environmental impact of machine learning is becoming an increasingly major issue for our field, given the extensive experimentation and hyper-parameter tuning required to successfully train large neural networks~\citep{strubell2019energy, schwartz_green_2019, lacoste_quantifying_2019}. In order to estimate our own carbon footprint, we counted all of the GPU hours (weighted by GPU usage) that our entire team used during the whole course of our project, from initial stages to model selection to the ablation study and hyper-parameter tuning on our final ClimateGAN model. We used the MLCO\textsubscript{2} Emissions Calculator\footnote{\href{https://mlco2.github.io/impact}{mlco2.github.io/impact}}~\citep{lacoste_quantifying_2019} to obtain a final estimate of \textbf{362.72 kilograms of CO$_2$eq.}, which is comparable to 900 miles driven by an average passenger vehicle, 42\% of a US household's yearly energy consumption~\footnote{Source: \href{https://www.epa.gov/energy/greenhouse-gas-equivalencies-calculator}{EPA}} or round trips between Paris, France and Saint-Petersburg, Russia (369kg CO$_2$eq.) or between Boston, MA and Miami, FL (345kg CO$_2$eq)~\footnote{Source: \href{https://www.icao.int/environmental-protection/Carbonoffset/Pages/default.aspx}{ICAO}}. This does not include the rest of the computational infrastructure's energy consumption (CPUs, data transfers and storage \textit{etc.})  nor the full Life Cycle Analysis of any of the hardware used.

We are incredibly lucky that our power grid is powered predominantly by renewable energy. If this were not the case and we were using coal-powered energy, the figure stated above would be 50-80 times larger, and thereby more problematic. We hope that our colleagues will also start tallying and sharing the carbon footprint of their research using tools such as \href{https://codecarbon.io/}{CodeCarbon} and the \href{https://github.com/Breakend/experiment-impact-tracker}{Experiment Impact Tracker} and that our community will start being more mindful regarding the trade-off between scientific progress and environmental impact.

\section{Full ClimateGAN Architecture}

In order to better illustrate the overall training procedure of ClimateGAN, we provide a more detailed overview in \cref{fig:full-climategan}. 

The Masker can be seen in the top part of the Figure, encompassing the three decoders described in~\cref{sec:masker}: Depth (D) , Segmentation (S) and Flood-Mask (M).  The Painter is shown in the lower part of the image, with its SPADE-based encoder P. The losses of each component are indicated in the rounded white boxes within each decoder. Tensors are represented using squares, with different colors for input, label, intermediate and output tensors, and pseudo-labels with dotted outlines. To emphasize the SPADE-based conditional architectures of M and P, they are colored in green. As per~\cref{fig:overview}, the output of ClimateGAN, \textit{i.e.} the flooded image $\tilde{\vy}$, is in blue. Note that the dotted and dashed line from $\vm_r$ to P is not leveraged during training and only used for test-time inferences. Discriminators are conceptually included in the WGAN losses.

\begin{figure}[!ht]
\centering
\includegraphics[height=0.91\textheight]{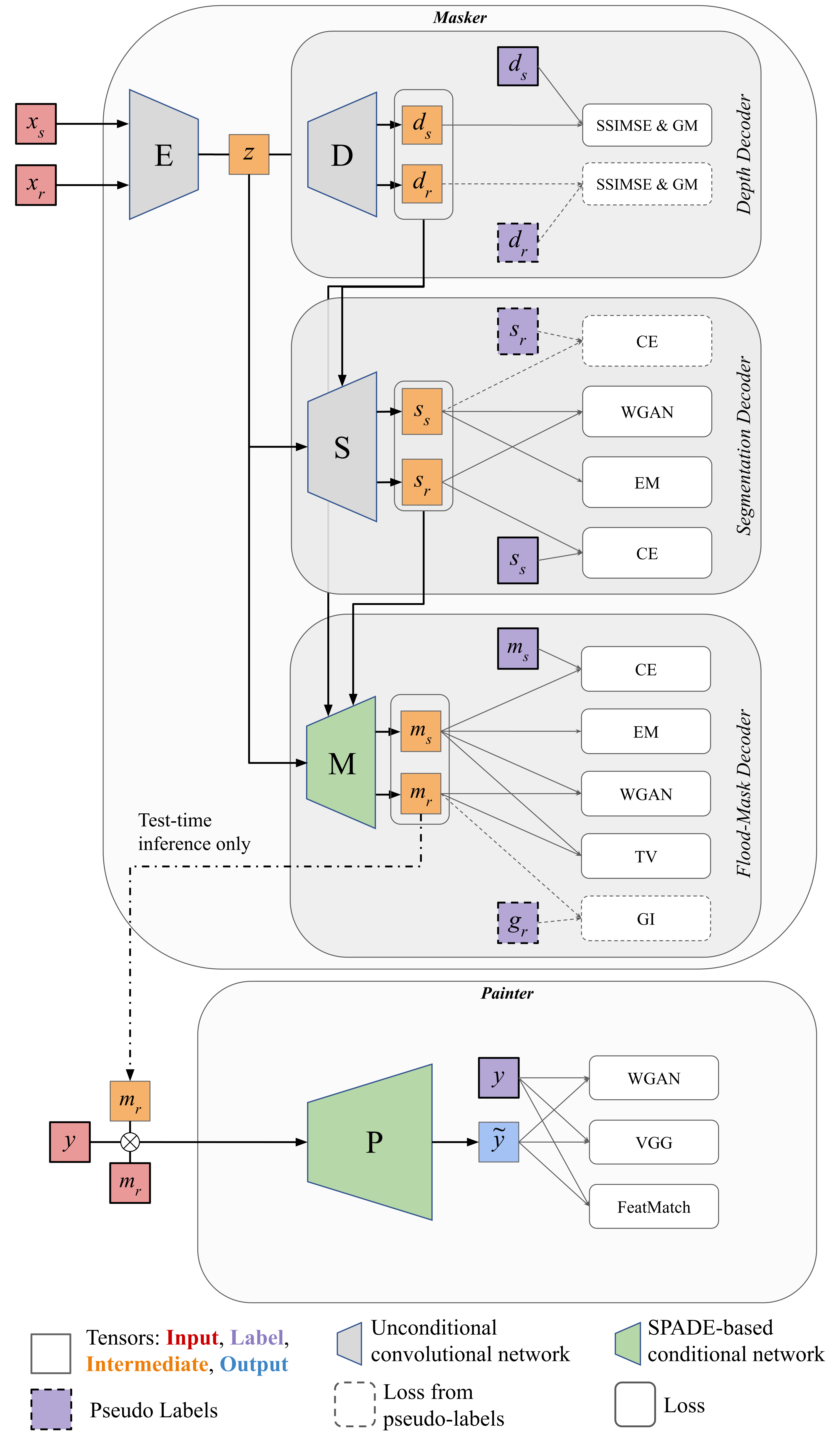}
\caption{Detailed diagram of the training procedure of ClimateGAN. We highlight how the simulated and real data paths in the model are similar yet different. We use dashed arrows and boxes to emphasize that pseudo labels are only used as noisy signal in the beginning of training. All the losses represented in the white rounded boxes are detailed in~\cref{sec:masker} and \cref{sup:losses}. }
\label{fig:full-climategan}
\end{figure}

\section{Supplementary Images} 
\label{sup:more-inferences}

In order to further illustrate the performance and capabilities of ClimateGAN, we provide additional inferences in \cref{fig:more-climategan}, complementing those presented in \cref{fig:climategan-inferences}. The first two rows were chosen to be successful inferences, middle two rows were selected at random and bottom two rows are failure cases.

\begin{figure}[ht!]
\small
\begin{tabularx}{\textwidth}{*{6}{Y}}
Input & Depth & Segmentation & Mask & Masked Input & Painted Input\\
\end{tabularx}
    \centering
    \includegraphics[width=0.975\textwidth]{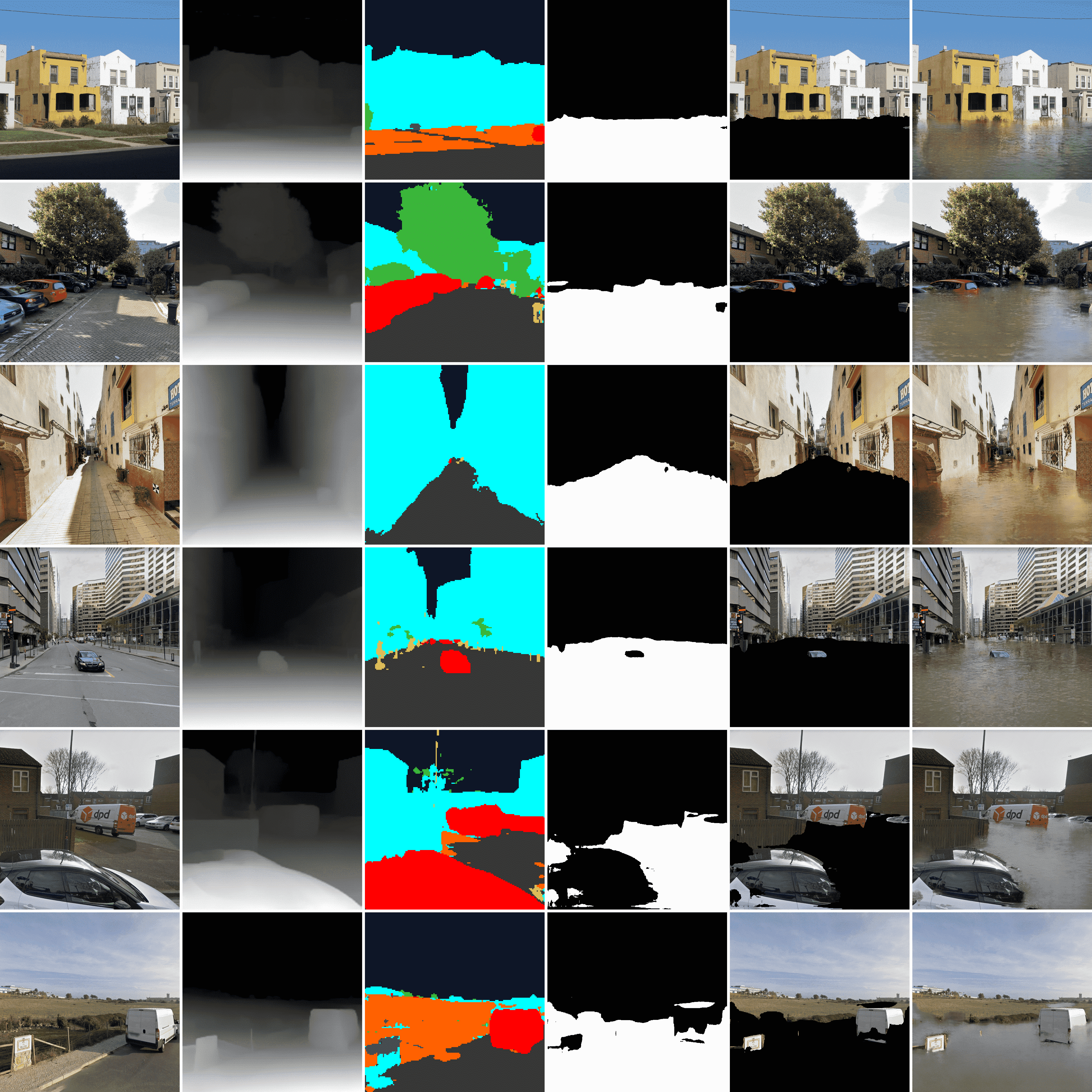}
    \vspace{0.2cm}
    \caption{More samples of the full ClimateGAN forward pass: input image, inferences from the depth, segmentation and flood mask decoders of the Masker, followed by the masked input image fed to the Painter and finally the flooded image output by the Painter. Notably, the Painter is able to produce consistently contextualized water, with color and reflections relevant to the surrounding objects and to the sky. While generally able to appropriately understand a scene's perspective and circumvent objects, the Masker is however sometimes unable to predict plausible flood masks (as illustrated in the bottom two rows). It is difficult to exactly understand the source of this because both the depth and segmentation maps look generally appropriate.}
    \label{fig:more-climategan}
\end{figure}

\begin{figure}[h!]
\centering
\includegraphics[height=0.9\textheight]{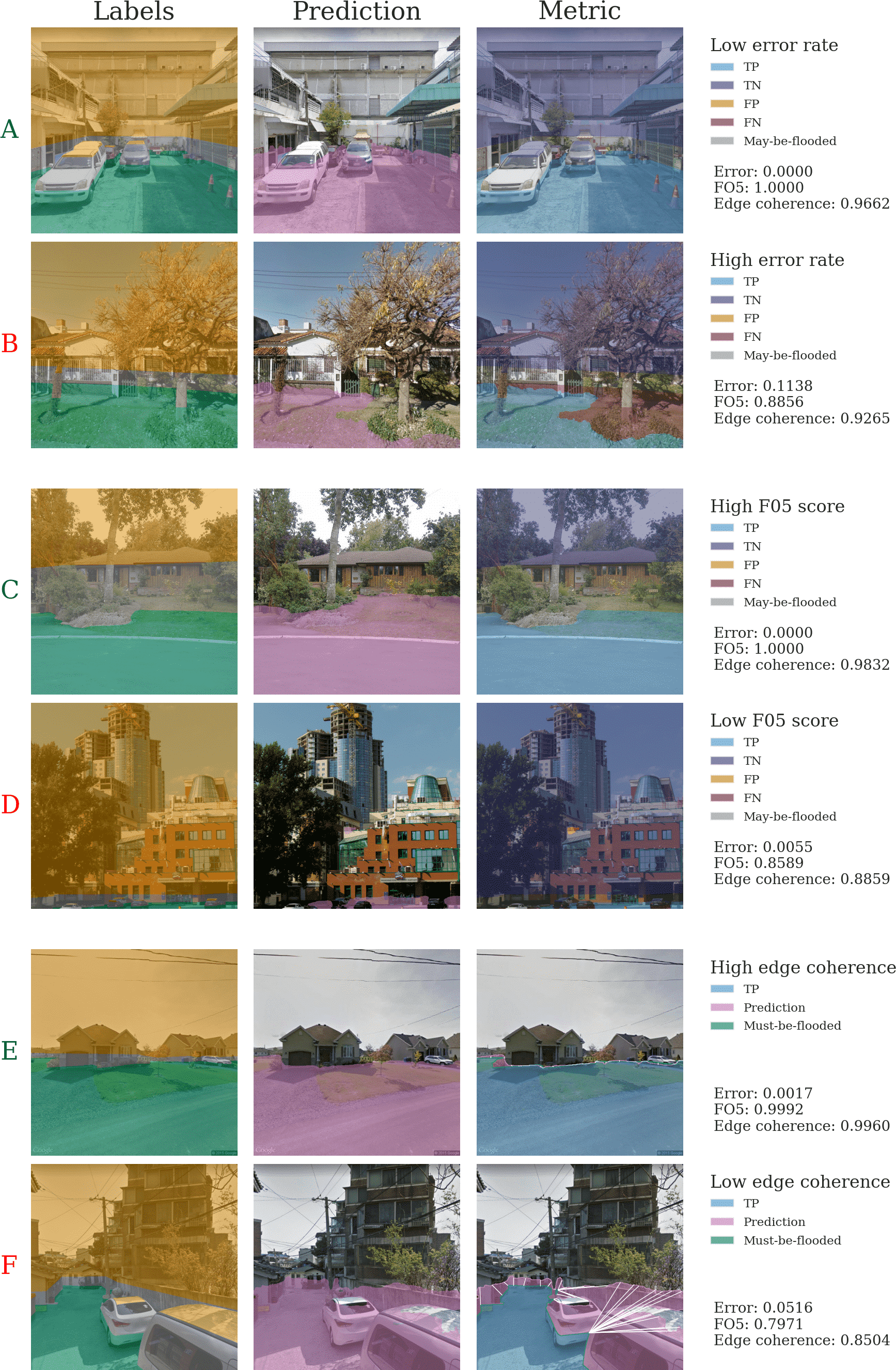}
\caption{Examples of images that obtained good and bad---within the 2nd and 98th quantiles, respectively---Masker's predictions metrics. From top two rows to bottom two: error, F05 score and edge coherence. The first row of each metric corresponds to an image with good values of the metric. The white segments in the images illustrating the edge coherence indicate the shortest distance between the predicted and the ground truth mask.    The legend of the column "Labels" is the same as in \cref{fig:labels}, and the images can be identified in \cref{fig:scatter_metrics} by the letters on the left.}
\label{fig:metrics-all}
\end{figure}

}

\end{document}